\documentclass[lettersize,journal]{IEEEtran}
\pdfoutput=1
\usepackage{amsmath,amsfonts}
\usepackage{bbm}
\usepackage{algorithmic}
\usepackage{algorithm}
\usepackage{array}
\usepackage[caption=false,font=scriptsize,labelfont=sf,textfont=sf]{subfig}
\usepackage{textcomp}
\usepackage{stfloats}
\usepackage{url}
\usepackage{verbatim}
\usepackage{graphicx}
\usepackage{cite}
\usepackage{booktabs}
\usepackage{multirow}
\usepackage{enumerate}
\usepackage{amsmath}
\usepackage{bigstrut}
\usepackage{color}

\newcommand{\tx}[1]{\textcolor[rgb]{0,0,0}{#1}}

\def\eg{{\it{e.g.}}}
\def\etal{{\it{et al.}}}
\def\ie{{\it{i.e.}}}

\begin{document}

\title{Boosting Night-time Scene Parsing with Learnable Frequency}
\author{Zhifeng Xie, Sen Wang, Ke Xu, Zhizhong Zhang, Xin Tan, Yuan Xie, Lizhuang Ma
\thanks{Zhifeng Xie is with the Department of Film and Television Engineering,
Shanghai University, Shanghai 200072, China, and also with Shanghai Engineering Research Center of Motion Picture Special Effects, Shanghai 200072, China.  E-mail: zhifeng\_xie@shu.edu.cn}
\thanks{Sen Wang is with the Department of Film and Television Engineering,
Shanghai University, Shanghai 200072, China. E-mail: wangsen@shu.edu.cn}
\thanks{Ke Xu is with the Department of Computer Science, City University of Hong Kong, HKSAR 999077, China. E-mail: kkangwing@gmail.com}
\thanks{Zhizhong Zhang, Xin Tan, Yuan Xie, and Lizhuang Ma are with the School of Computer Science and Technology, East China Normal University, Shanghai, China. Lizhuang Ma is also with the Department of Computer Science and Engineering, Shanghai Jiao Tong University, Shanghai 200240, China. E-mail: zzzhang@cs.ecnu.edu.cn, xtan@cs.ecnu.cn, xieyuan8589@foxmail.com, lzma@cs.ecnu.edu.cn}
\thanks{Manuscript received xx xx, 2022; revised xx xx, 2022}}


\markboth{Journal of \LaTeX\ Class Files,~Vol.~14, No.~8, August~2021}%
{Shell \MakeLowercase{\textit{et al.}}: A Sample Article Using IEEEtran.cls for IEEE Journals}

\IEEEpubid{}

\maketitle

\begin{abstract}
Night-Time Scene Parsing (NTSP) is essential to many vision applications, especially for autonomous driving. Most of the existing methods are proposed for day-time scene parsing. They rely on modeling pixel intensity-based spatial contextual cues under even illumination. Hence, these methods do not perform well in night-time scenes as such spatial contextual cues are buried in the over-/under-exposed regions in night-time scenes.
In this paper, we first conduct an image frequency-based statistical experiment to interpret the day-time and night-time scene discrepancies. We find that image frequency distributions differ significantly between day-time and night-time scenes, and understanding such frequency distributions is critical to NTSP problem. Based on this, we propose to exploit the image frequency distributions for night-time scene parsing. 
First, we propose a Learnable Frequency Encoder (LFE) to model the relationship between different frequency coefficients to measure all frequency components dynamically.
Second, we propose a Spatial Frequency Fusion module (SFF) that fuses both spatial and frequency information to guide the extraction of spatial context features.
Extensive experiments show that our method performs favorably against the state-of-the-art methods on the NightCity, NightCity+ and BDD100K-night datasets. In addition, we demonstrate that our method can be applied to existing day-time scene parsing methods and boost their performance on night-time scenes.

\end{abstract}


\section{Introduction}

\begin{figure}[!t]
    \centering
    \subfloat[\rmfamily{Examples of Cityscapes and NightCity}]{
            \centering
          \includegraphics[width=0.9\linewidth]{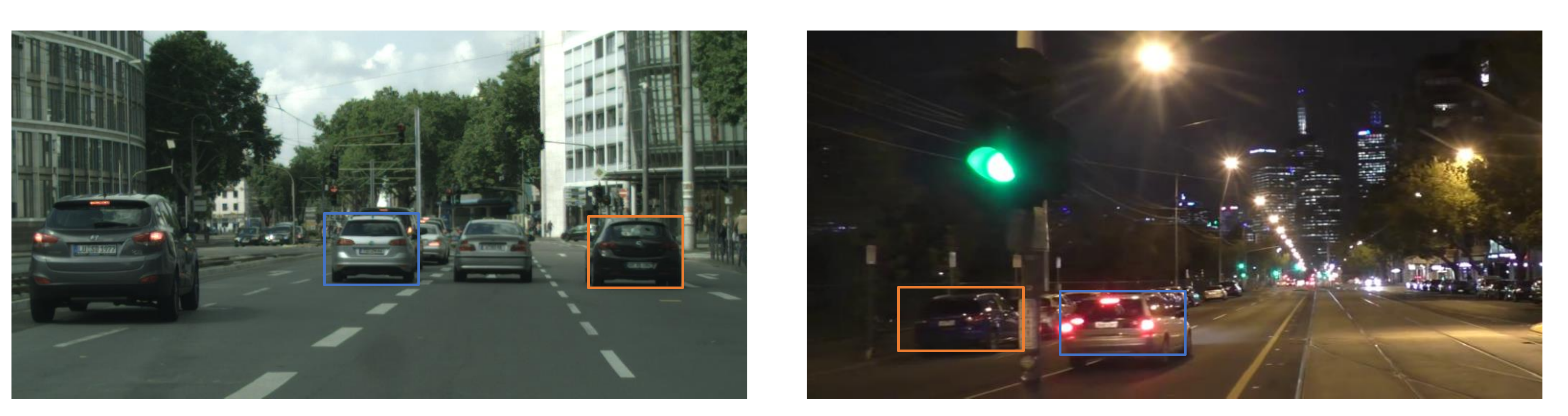}
          \label{fig:intro_1_1}
    }
    \qquad 
    \vspace{1pt}
    \subfloat[\rmfamily{Image-level Frequency Analysis}]{
           \centering
           \includegraphics[width=0.9\linewidth]{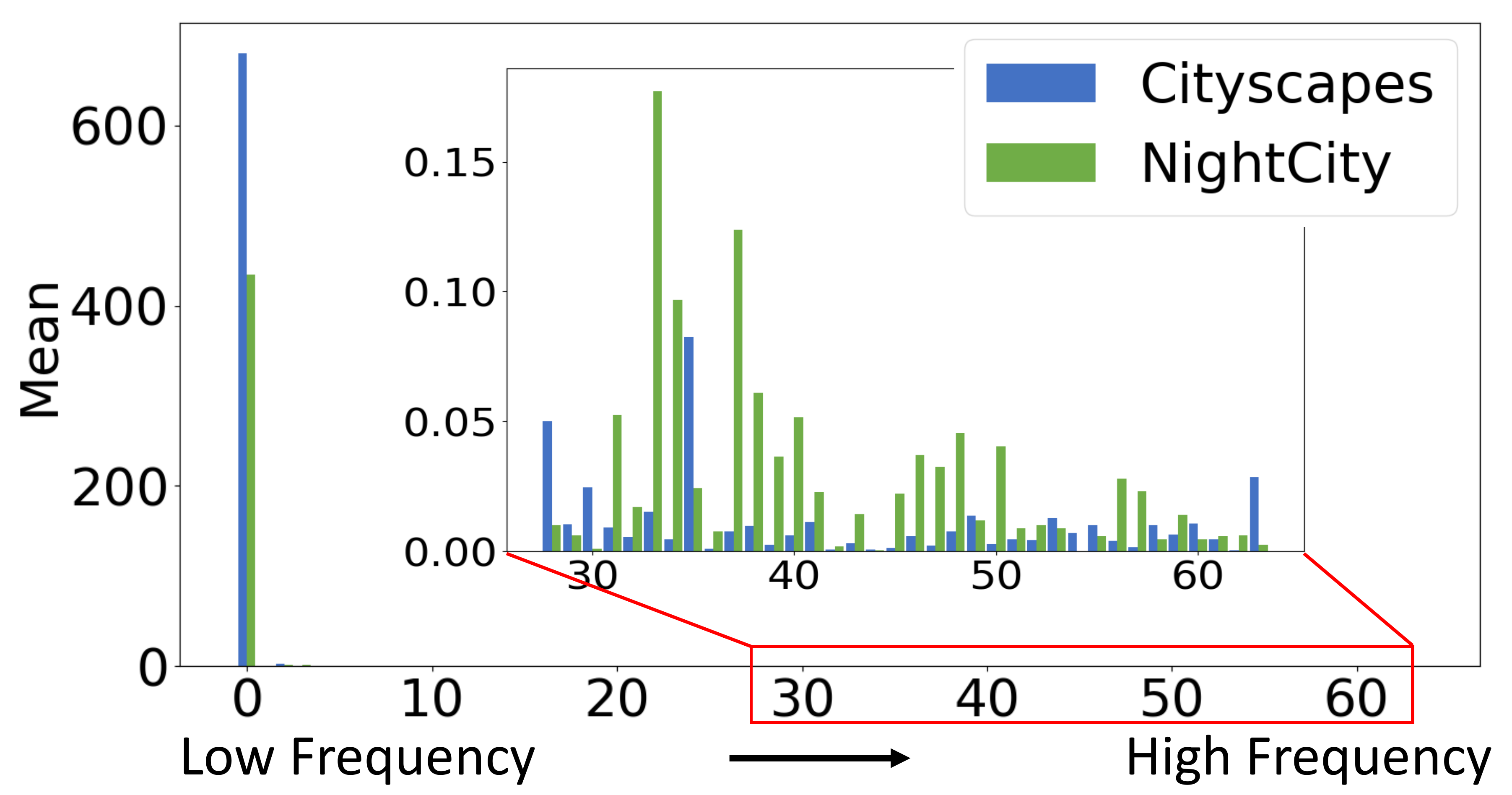}
            \label{fig:intro_1_a}
            }
    \qquad        
    \vspace{1pt}
    \subfloat[\rmfamily{Local Under-/Over-exposures Frequency Analysis}]{
           \centering
           \includegraphics[width=0.92\linewidth]{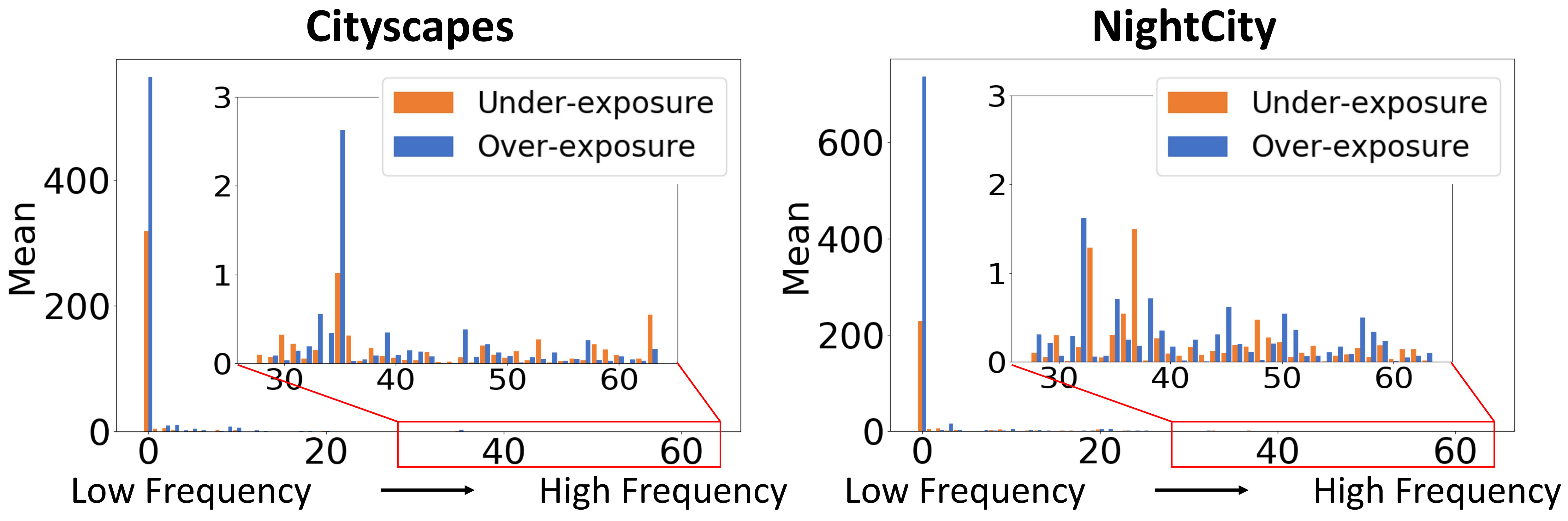}
            \label{fig:intro_1_b}
            }   

\caption{Image-level frequency statistics. 
(a) shows one day-time scene from Cityscapes (left) and one night-time scene from NightCity (right).
(b) shows image-level frequency distributions of two images of (a). (c) shows local frequency distribution of regions with under-/over-exposures (marked with orange and blue boxes, respectively) of images in (a). The high-frequency components are zoomed in by red box.}
\label{fig:intro_1}
\end{figure}

\IEEEPARstart{S}{cene} parsing is a fundamental task in computer vision with many downstream applications, such as autonomous driving\cite{fujiyoshi2019deep}, human parsing\cite{li2020self}, and image inpainting\cite{liao2021image}. 
Most representative scene parsing methods~\cite{long2015fully, zhao2017pyramid, chen2018encoder, huang2019ccnet, zheng2021rethinking} are proposed for day-time scenes. However, while night-time may contribute to half of total working hours (\eg, in autonomous driving), these existing methods do not work well in night-time scenes due to the day-time/night-time scene discrepancies (see Figure~\ref{fig:intro_1_1}). Meanwhile, although there are some methods~\cite{dai2018dark, sakaridis2019guided, wu2021dannet, xu2021cdada,yang2020fda} proposed to transfer the day-time domain knowledge to the nigh-time domain for scene parsing through domain adaptation, they still cannot achieve practical performances due to the less resolved domain discrepancies.

Recently, Tan~\etal~\cite{tan2021night} propose the first large-scale night-time scene dataset (NightCity). They also propose an exposure-guided network for night-time scene parsing (NTSP). Deng~\etal~\cite{deng2022nightlab} propose the NightLab, which further boosts the performance of NTSP by learning the image lighting variation and mining hard segmented regions.

However, all these methods typically rely on modeling pixel-intensity-based contextual features, which are not necessarily reliable under uneven night-time lighting conditions.
On the other hand, we note that some style transfer-based segmentation methods~\cite{yang2020fda,xu2021cdada} assume that the low-level spectrum represents scene lighting information.
Hence, two questions are raised: {\it Can image frequency distributions represent the day-time/night-time domain discrepancies? And are all frequency components are important for NTSP?} 

To answer the aforementioned two questions, we first conduct an image-frequency based analysis. We first analyze image-level frequency distributions by randomly select one day-time image from the Cityscapes\cite{cordts2016cityscapes} and one night-time image from the NightCity~\cite{tan2021night} (Figure \ref{fig:intro_1_1}).
We use the Discrete Cosine Transform (DCT) to compute the spectrogram of images as in \cite{xu2020learning}.
Following the JPEG compression process\cite{wallace1992jpeg}, the image is divided into multiple 8 × 8 blocks. 
Then, we calculate the mean {value} of spectrograms of all blocks as shown in Figure \ref{fig:intro_1_a}.
While the frequency distribution of day-time image does differ from that of night-time image and such difference mainly comes from the low frequency components, we can see that night-time images do have different high frequency distribution from day-time image.

We further analyze the local regions of the night-time image where under- and over-exposures happen (marked with orange and blue boxes Figure \ref{fig:intro_1_1}). For the corresponding comparing regions of day-time image we select the objects with the same semantics (\ie, cars). We compute the spectrograms of those regions as shown in Figure \ref{fig:intro_1_b}. We can see that the high frequency distribution of day-time image tends to have less peaks due to its relatively even lighting condition, while that of night-time image tends to have more peaks. This demonstrates that high frequency distribution differences reveal the lighting discrepancies of different domains.

\begin{figure}[!t]
    \centering
    \subfloat[\rmfamily{Spectrogram Segmentation}]{
           \centering
           \includegraphics[width=0.36\linewidth]{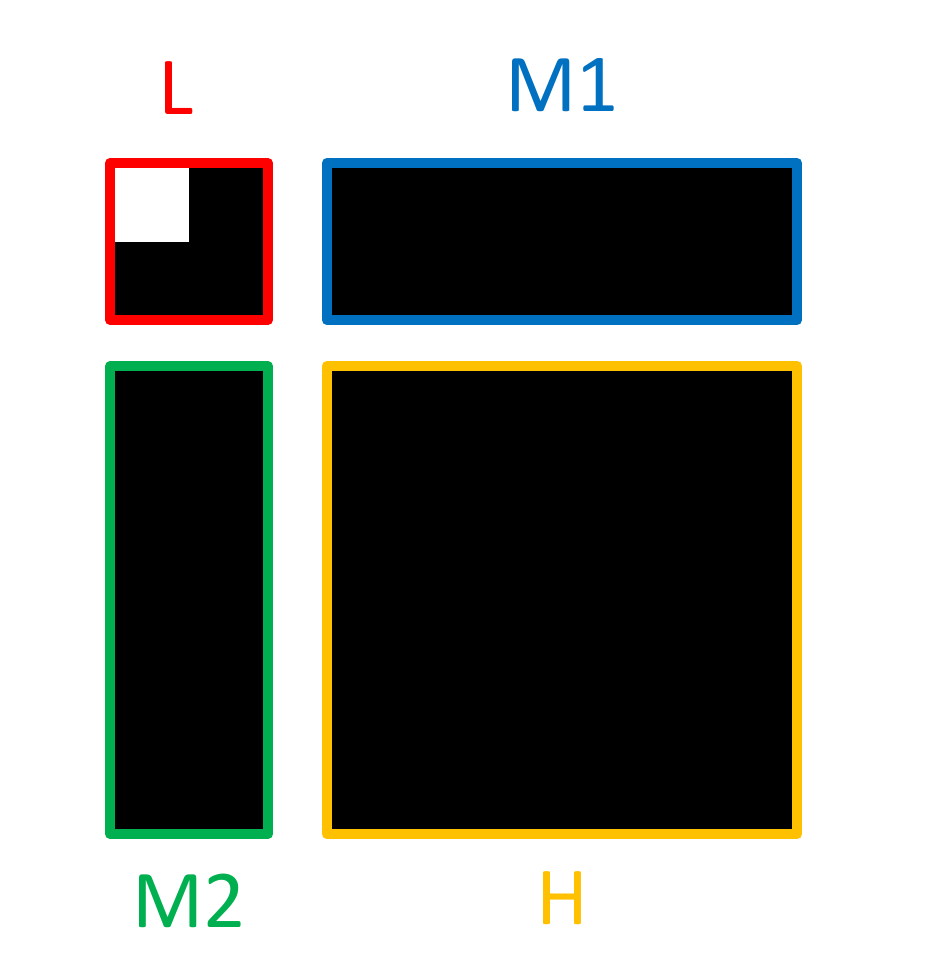}
            \label{fig:intro_2_c}
            }   
    \subfloat[\rmfamily{Dataset-level Frequency Analysis}]{
           \centering
           \includegraphics[width=0.56\linewidth]{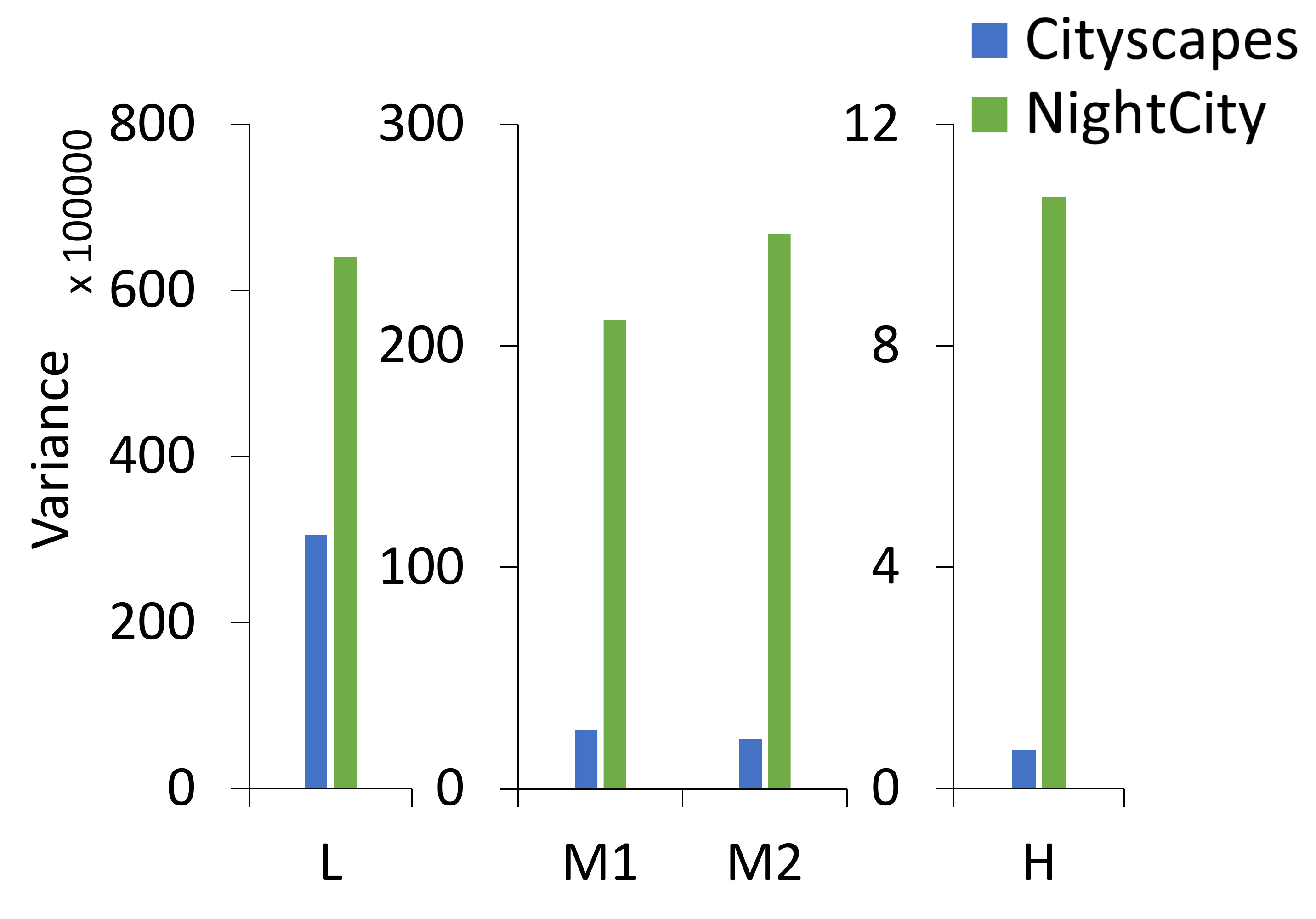}
            \label{fig:intro_2_d}
            }
    \caption{Dataset-level frequency statistics. (a) The spectrogram is divided into four parts, the low frequency (L), mid frequency (M1, M2) and high frequency (H), in which the low frequency area occupies 1/16 of the frequency map, the two mid frequency areas occupies 3/16, and the high frequency area occupies 9/16. (b) Dataset-level mean variance statistics across the four frequency regions.}
    \label{fig:intro_2}
\end{figure}

Furthermore, we perform quantitative experiments at the dataset level to demonstrate our observation. 
To better analyze the frequency difference, we divide the spectrogram into four parts, as shown in Figure \ref{fig:intro_2_c}.
First, we calculate the mean \tx{values} of the spectrogram in each frequency region, and then calculate the variance of the mean \tx{values} of each frequency region of all night images in the dataset separately. 
We show the results in Figure \ref{fig:intro_2_d} that the variance of the night-time scenes in each region is larger than that of the day-time scenes, which indicates that the difference of the dataset-level frequency information for night-time scenes is also significant.
This motivates us to design a network for NTSP that is learnable for all frequency information \tx{to adjust the frequency components dynamically.}

In this paper, we propose a novel Frequency Domain Learning Network (FDLNet), which first \tx{deals} with the NTSP in the frequency domain. 
Specifically, we first propose a Learnable Frequency Encoder (LFE) which fully exploits all frequency components generated by DCT to adjust the channel response of different frequency components dynamically. 
Since the frequency distribution of different night-time images is diverse, the LFE can adaptively adjust the channel response of frequency components, so the weights of frequency \tx{components} are unique for each image. 
Then, we propose a Spatial Frequency Fusion module (SFF), which fuses the spatial features and frequency features in channel-wise.
We use both spatial and frequency information to guide the extraction of spatial context features for NTSP.
%
%
We conduct extensive experiments on night-time datasets (including NightCity, NightCity+ and BDD100K-night), showing that our method plays favorably against state-of-the-art methods. Besides, our method can be easily applied to existing state-of-the-art day-time segmentation methods\cite{zhao2017pyramid} \cite{chen2018encoder} \cite{huang2019ccnet} to adapt them for NTSP.

In sum, our main contributions are:
\begin{enumerate}
\item{We interpret the scene lighting discrepancies between day-time and night-time scenes with the image frequency distributions. We show that a full understanding of image frequency distributions is crucial to NTSP. We propose a novel Frequency Domain Learning Network (FDLNet) for NTSP.}


\item{We propose the Learnable Frequency Encoder (LFE), to dynamically adjust the channel responses of all frequency components. We propose the Spatial Frequency Fusion module (SFF), which leverages the frequency information to model spatial contexts by a fusion of spatial and frequency features.}
\item{We show that our method can easily be applied to state-of-the-art day-time scene parsing methods to boost their performances for NTSP.}
\end{enumerate}

\section{Related Work}
\subsection{Scene Parsing}
Scene parsing aims to assign each pixel with its class label. Long~\etal~\cite{long2015fully} propose the first fully convolutional network (FCN) to extract deep features for scene parsing.
Later, many methods such as PSPnet\cite{zhao2017pyramid} and Deeplab\cite{chen2017deeplab} \cite{chen2017rethinking} \cite{chen2018encoder} are proposed to aggregate more spatial features by expanding their reception fields.
%
In order to obtain more effective spatial features, a variety of attention mechanisms have been studied in scene parsing.
In~\cite{zhao2018psanet}, Point-wise Spatial Attention is proposed to associate the information of each location with that of other locations.
Self-attention mechanism is introduced in DANet\cite{fu2019dual} and OCNet\cite{yuan2018ocnet} to capture contextual information.
CCnet\cite{huang2019ccnet} and Axial-DeepLab\cite{wang2020axial} apply the Non-local module\cite{wang2018non} to model long-range spatial contextual information.
Recently, transformer-based methods are proposed to model global contexts for scene parsing.
STER~\cite{zheng2021rethinking} uses the transformer layers to form the encoder for extracting the global context information.
Swin Transformer\cite{liu2021swin} uses the sliding windows with information exchange mechanism to reduce the computational complexity of transformer, while capturing global information.
Strudel~\etal~\cite{strudel2021segmenter} propose a fully transformer architecture with a Mask Transformer as the decoder to generate class masks. Xie~\etal~\cite{xie2021segformer} propose to fuse multi-level features without positional encoding in the encoding stage.

Meanwhile, there are also some methods proposed to encode prior knowledge into scene parsing.
HANet~\cite{choi2020cars} models the height distribution statistics of object categories, based on which they propose to learn height-driven attention.
SANet~\cite{zhong2020squeeze} factorizes the scene parsing task into two sub-tasks of pixel classification and pixel grouping, {and leverages pixel grouping to aggregate contextual information to enhance pixel classification.}
{ISNet~\cite{jin2021isnet} learns both image level and semantic level contextual features to model inter- and intra-class correlation for scene parsing.
STLNet~\cite{zhu2021learning} uses the proposed Quantization and Counting Operator to leverage the low-level texture features for scene parsing.} 
In~\cite{jin2021mining}, contextual information beyond a single image is modeled via their proposed MCIBI by dynamically building dataset-level semantic features during training.

All above-mentioned methods are proposed for day-time scene parsing.
Due to the lack of large-scale night-time datasets, previous NTSP methods have to resort to semi-supervised learning~\cite{feng2022dmt} or domain adaption~\cite{dai2018dark,sakaridis2019guided,wu2021dannet,xu2021cdada}, which cannot address the domain discrepancies between day-time and night-time scenes.
Most recently, Tan~\etal~\cite{tan2021night} propose a large-scale real night-time dataset and an exposure-guided network to learn robust semantic features. Deng~\etal~\cite{deng2022nightlab} propose the NightLab, which further boosts the performance of NTSP by learning the image lighting variation and mining hard segmented regions.

All previous methods rely on pixel-intensity based spatial contextual information, which may not be reliable due to the existence of over- and under-exposures in night-time scenes. In this paper, we study the NTSP problem from the image frequency perspective, showing that an understanding of frequency distributions facilitates contextual information modeling significantly.

\subsection{Deep Learning in the Frequency Domain}
Deep learning in the image frequency domain has many applications of, \eg, image restoration~\cite{liu2018multi} and demoireing~\cite{zheng2021learning}, model compression~\cite{chen2016compressing}, image classification~\cite{xu2020learning,gueguen2018faster,Ehrlich_2019_ICCV} and instance segmentation~\cite{qin2021fcanet}. Their main idea is to select a set of (low-frequency) components to reduce the network computational complexity.
To reduce the high-frequency information loss of the downsampling process, a content-aware anti-aliasing module is proposed in \cite{zou2020delving}.
In \cite{xu2020learning}, the Discrete Cosine Transform (DCT) is used to preprocess the image to reduce the loss of important information in the process of downsampling.

Particularly in scene parsing, previous methods~\cite{ding2019boundary, takikawa2019gated, yuan2020segfix, ma2021boundary, borse2021inverseform} mainly focus on the image boundary information in the gradient domain.
~\cite{li2020improving} decouples the body (low frequency) and edge (high frequency) features of the image to optimize the boundary details of the prediction results.
In~\cite{yang2020fda}, style transfer from day-time to night-time is performed in the Fourier domain. However, their assumption of the low-frequency image amplitude component representing the whole scene illumination does not always hold true (\eg, when both over- and under-exposure happen).

Different from previous work, we propose to model the whole image frequency distributions and combine them with pixel intensity-based contextual features for NTSP.

\section{The Proposed Method}

\subsection{Overview}
\label{sec::overview}

\begin{figure*}[!t]
\centering
\includegraphics[width=0.96\textwidth]{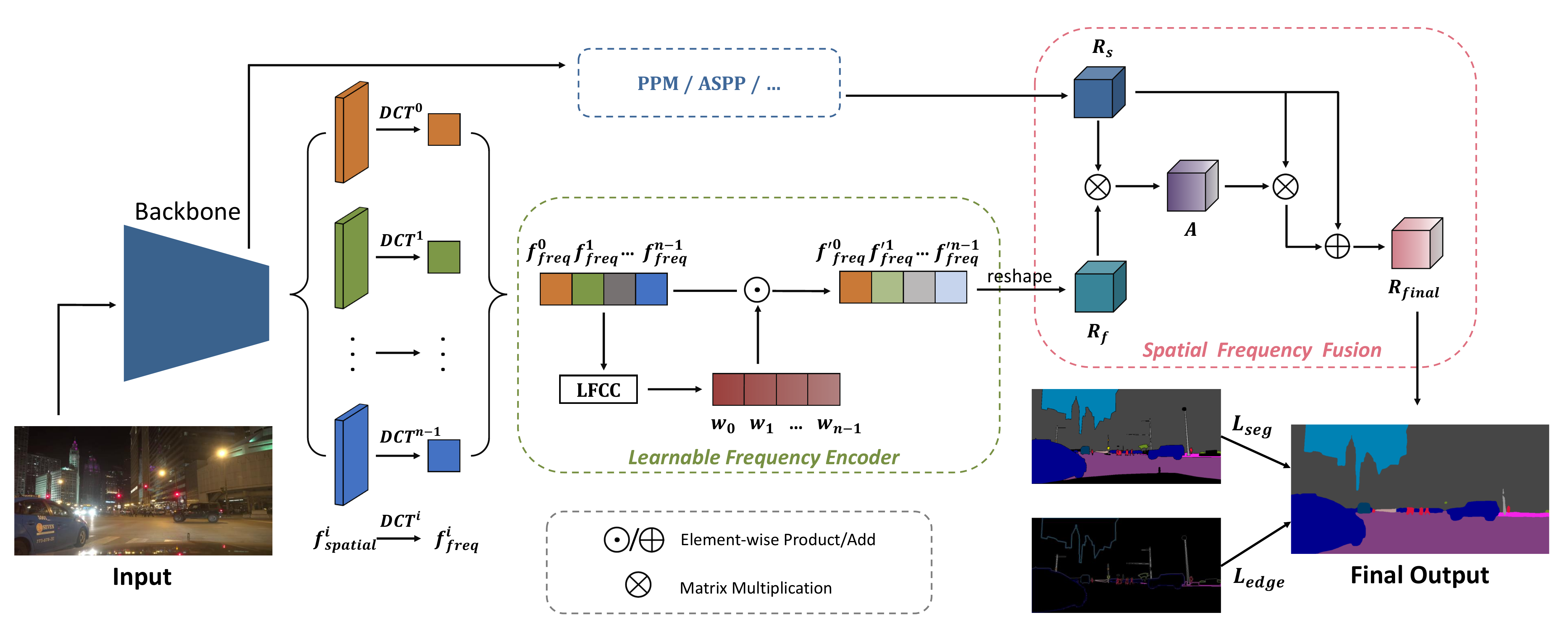}
\caption{Illustrating the pipeline of FDLNet. 
Given an input image, the backbone extracts spatial features. 
We leverage discrete cosine transform (DCT) in the network to get the frequency features from the output of the backbone. 
Then, we propose a Learnable Frequency Encoder (LFE) to leverage learnable frequency to guide the network to segment. 
To this end, a Learnable Frequency Component Convolutional layer (LFCC) is proposed to dynamically adjust the weights of all frequency components and we reshape it to obtain the frequency representations $R_{f}$. 
Meanwhile, we leverage existing spatial context aggregation modules (\eg{}PPM\cite{zhao2017pyramid}, ASPP\cite{chen2018encoder}) to extract the spatial representations $R_{s}$. 
We feed both $R_{f}$ and $R_{s}$ to the Spatial Frequency Fusion module (SFF) to obtain the affinity representations $A$.
The $A$ guides the $R_{s}$ as an attention map which
adjusts the channel response to get the final representations $R_{final}$.
Finally, we utilize a segmentation head to generate the prediction from the fused feature map.
Best viewed in color.}
\label{fig:pipeline}
\end{figure*}

In this paper, we present a novel method that models frequency distribution to facilitate the night-time scene parsing.
Figure~\ref{fig:pipeline} shows the pipeline of the proposed method. Given an RGB image $I$, the backbone encodes the images into a spatial feature map, denoted as $f_{spatial}$.
Then, we compute frequency features $f_{freq}$ by transforming $f_{spatial}$ into the frequency domain with Discrete Cosine Transform. 
To fully exploit frequency information, we first propose a Learnable Frequency Encoder (LFE). 
This module re-weights frequency feature $f_{freq}$ based on the contribution of each frequency component. 
{Second, we propose a novel spatial frequency fusion module to fuse the spatial $f_{spatial}$ and frequency $f_{freq}$ information in channel-wise.}
After fusion, a standard segmentation head is attached to produce the final parsing results.

\subsection{2D Discrete Cosine Transform}
\label{sec::dct}
We employ Discrete Cosine Transform (DCT) to transfer the spatial feature map to frequency domain. 
First, we simply review the principle of DCT.  
The basic function of the two-dimensional discrete cosine transform $B$ is:
\begin{equation}
\label{eq:DCT_1}
B_{x, y}^{u, v}=\cos \frac{(2x+1)u \pi}{2 N} \cos \frac{(2y+1)v \pi}{2 N} \text {, }
\end{equation}
where $u$ and $v$ are the horizontal and vertical frequency components, respectively. {$N$ is the size of an image block, and (x, y) represents the spatial locations of the image block}.
Then the two-dimensional discrete cosine transform can be formulated as:
\begin{equation}
\label{DCT}
F(u, v)=c(u) c(v) \sum_{x=0}^{N-1} \sum_{y=0}^{N-1} f(x, y) B_{x, y}^{u, v} \text {, }
\end{equation}
where $F(u, v)$ is the 2D DCT frequency spectrum, $u,v\in \{0,1, \cdots, n-1\}$, and $f(x, y)$ is a two-dimensional vector element of $N \times N$ in the spatial domain, $x,y\in \{0,1, \cdots, N-1\}$. $c(u)$ and $c(v)$ are compensation factors, written as:

\begin{equation}
\label{eq:DCT_2}
c(u), c(v) =\left\{\begin{array}{ll}
\sqrt{\frac{1}{N}}, & u,v=0 \\
\sqrt{\frac{2}{N}}, & u,v \neq 0 \text {. }
\end{array}\right.
\end{equation}

{Following \cite{qin2021fcanet}}, we utilize DCT to record the frequency information in the channel dimension.
Given input spatial feature map $f_{spatial} \in \mathbb{R}^{C \times H \times W}$, where $C, H$ and $W$ denote the channel dimension, height and width, respectively. 
According to the rules of image compression and coding, we reconstruct the size of $f_{spatial}$  into $N \times N$.
{Then, $f_{spatial}$ is divided into multiple parts in the channel dimension to obtain $f_{spatial}^{i} \in \mathbb{R}^{\frac{C}{n} \times H \times W}$, where $n$ is the total number of frequency components.}
Thus, we can obtain each frequency component $f_{freq}^{i}$ by its corresponding spatial feature component $f_{spatial}^{i}$ using 2D DCT function $\operatorname{DCT^i}$:
\begin{align}
\label{eq:DCT_4}
\nonumber f_{freq}^{i} &=\operatorname{DCT^i}\left( f_{spatial}^{i}(x, y) \right)\\
\nonumber &= c(u) c(v) \sum_{x=0}^{N-1} \sum_{y=0}^{N-1} f_{spatial}^{i}(x, y) B_{x, y}^{u, v}\\ 
&\text { s.t. } i \in\{0,1, \cdots, n-1\} \text {. }
\end{align}
After that, the multi-spectral frequencies vector $V_{freq}\in \mathbb{R}^{C}$ is defined as:
\begin{equation}
\label{eq:DCT_3}
V_{freq} =\operatorname{cat}\left(\left[{f_{freq}^{0}}, {f_{freq}^{1}}, \cdots, {f_{freq} ^{n-1}}\right]\right) \text{, }
\end{equation}
where $cat$ denotes concatenate operation.


\subsection{Learnable Frequency Encoder}
\label{sec::dffa}

Unlike day-time scenes, the frequency distribution of night images is more discrete (see Figure \ref{fig:intro_1} and \ref{fig:intro_2}). 
Simply using a fixed number of frequency components cannot handle night-time scene parsing. Hence, we propose the Learnable Frequency Encoder (LFE) to learn the importance of each frequency component.
{To dynamically adjust each frequency component, a Learnable Frequency Component Convolutional layer (LFCC)
{is used to convert the entire multi-spectral frequency vector $V_{freq}$ into the weight of each frequency component $W$, as:}
\begin{equation}
\begin{aligned}
\label{eq:FF_3}
W = softmax\left(LFCC \left(V_{freq}\right)\right) \text{, }
\end{aligned}
\end{equation}
where  $LFCC$ includes a $1 \times 1$ convolutional layer and a batchnorm layer. For training stability, we constrain the weights of LFCC to be positive and sum them to 1 by a $softmax$ function.
The $V_{freq}$ is reshaped to the size of $n \times \frac{C}{n} \times 1 \times 1$ and $W \in \mathbb{R}^{n \times 1^2 \times 1 \times 1}$, where each weight of the $1^2$ channel corresponds to one frequency component $f_{freq}^{i}  \in \mathbb{R}^{\frac{C}{n} \times 1 \times 1}$.} 
This operation can be expressed as:
\begin{equation}
\begin{aligned}
\label{eq:FF_4}
f_{freq}^{' i} =  w^{i} \cdot f_{freq}^{i} \text{, }
\end{aligned}
\end{equation}
where $w^{i}$ is one channel of $W$ corresponding to each frequency component $f_{freq}^{i}$.
Then we calculate the re-weight multi-spectral frequencies vector $V_{freq}^{'}$ as follows:
\begin{align}
\label{eq:FF_5}
\nonumber V_{freq}^{'} &=W V_{freq} \\
\nonumber &=\operatorname{cat}\left({ w }^{0} { f_{freq}^{0}}, { w }^{1}{ f_{freq}^{1}}, \cdots, { w }^{n-1}{ f_{freq}^{n-1}}\right)\\
&=\operatorname{cat}\left({ f_{freq}^{' 0}}, { f_{freq}^{' 1}}, \cdots, { f_{freq}^{' n-1}}\right) \text{. }
\end{align}
We use $n$ to group the consecutive channels of frequency vector $V_{freq}$ and the output of filter $W$ adjusts the weight of each frequency component $f_{freq}^{i}$ based on $V_{freq}$. 
By multiplying $w$ and $V_{freq}$ element-wise, the encoder is learnable to predict the weight of each frequency component. 
Finally, the output of the encoder is the re-weight frequency feature $V_{freq}^{'}$ which is rectified at the channel dimension. 

\textit{Discussion:} We note that there are some methods~\cite{xu2020learning,qin2021fcanet} proposed to model image frequency information but only select top $k$ frequency components to represent the whole image. However, as shown in Figure \ref{fig:intro_1} and \ref{fig:intro_2}, high-frequency components still contain important information due to the uneven lighting conditions of night-time scenes. Our module models the whole image frequency distribution and adjusts their weights dynamically. The experiment in Table~\ref{tab:ab_2} shows that dynamically modeling the whole image frequency distribution facilitates the NTSP performance.

\subsection{Spatial Frequency Fusion}
\label{sec::fusion}

Modeling the frequency distribution helps the network understand the scene illumination. We then use the learned frequency features to guide the network to model spatial context features for night-time scene parsing.
Specifically, we propose the Spatial Frequency Fusion module (SFF) to fuse features from two different domains. 
First, we employ a spatial context aggregation module to enhance the extraction of spatial features $f_{spatial}$, and then utilize a convolutional layer to transform the $f_{spatial}$ into spatial representations ${R}_{s}\in \mathbb{R}^{C\times \frac{H}{8} \times \frac{W}{8}}$. 
Meanwhile, the re-weight frequency feature $V_{freq}^{'}$ is fed into a convolution layer to reduce the dimensionality to generate frequency representations ${R}_{f}$.
After that ${R}_{f}\in \mathbb{R}^{C\times 1 \times 1}$ extended to ${R}_{f}\in \mathbb{R}^{C\times \frac{H}{8} \times \frac{W}{8}}$ so as to keep the same shape with $R_s$.
Then both ${R}_{f}$ and ${R}_{s}$ are reshaped to $\mathbb{R}^{C\times D}$, where $D=\frac{H}{8} \times \frac{W}{8}$. 
We conduct matrix multiplication between the transpose of reshaped ${R}_{f}$ and ${R}_{s}$, and apply a $Softmax$ layer to calculate the affinity map.
The affinity operation is then defined as:

\begin{equation}
\label{eq:SFCF_1}
A (i, j)= \frac{\exp \left(R_{s}^i \otimes \left( R_{f}^j \right) ^{T}\right)}{\sum_{i=1}^{C} \exp \left(R_{s}^i \otimes \left (R_{f}^j \right)^{T} \right)} \text{, }
\end{equation}
where $A (i, j)$ indicates the effect of $i^{th}$ channel in the spatial representations ${R}_{s}$ on the $j^{th}$ channel in the frequency representations ${R}_{f}$ and $\otimes$ denotes matrix multiplication.
$A$ is the affinity map calculated over the channel dimension.
After that, the final fused representation ${R}_{final}$ is calculated as follows:
\begin{equation}
\label{eq:SFCF_2}
R_{final}=\alpha \left(permute \left(A \otimes R_{s} \right)\right) + R_{s} \text{, }
\end{equation}
where $permute$ reshapes the result of $A \otimes R_{s}$ to $C \times \frac{H}{8} \times \frac{W}{8}$ and $\alpha$ is a scale parameter to reduce gradient instability. 
Note that each channel of $R_{final}$ is the weighted sum of all channels through spatial and frequency features, and effectively captures the long-term dependencies between spatial and frequency domains. 

\subsection{Loss Function}
\label{sec::overv loss}
We use the standard cross-entropy loss to measure the errors between the network predictions and ground truth labels.
In addition, since the high-frequency boundary information is an important cue for scene parsing, we also incorporate edge loss during training. Unlike previous methods~\cite{takikawa2019gated,li2020improving} that learn edge information in the spatial domain, which are not reliable in night-time scenes due to their complex lighting conditions, we propose to learn edge information in the frequency domain.


Let $s_{i, c}$ and $\hat{s_{i, c}}$ be the ground-truth and prediction results of the $i^{th}$ pixel of class $c$, respectively. 
$L_{edge}$ focus on the semantic edge regions of $s_{i, c}$ as:
\begin{equation}
\label{eq:L_1}
L_{edge}=-\sum_{i} \sum_{c} \mathbbm{1}_{b_{i}} \cdot\left(s_{i, c} \log \hat{s_{i, c}}\right) \text{, }
\end{equation}
where $L_{edge}$ represents cross-entropy loss on semantic edge regions. 
$b_{i}$ is the ground-truth semantic edge of the $i^{th}$ pixel and the $\mathbbm{1}_{b_{i}}$ represents indicator function that semantic edge region in the ground-truth $s_{i, c}$.

The overall loss $L$ can be defined as:
\begin{equation}
\label{eq:L_2}
L=\lambda_{1} L_{\text {seg }}+\lambda_{2} L_{\text {edge}} \text{, }
\end{equation}
where $L_{seg}$ is a standard cross-entropy loss. 
$\lambda_{1}$ and $\lambda_{2}$ are two hyperparameters that control the weighting between the losses.


\section{Experiments}
To evaluate our proposed method, we conduct extensive experiments on NightCity \cite{tan2021night}, NightCity+ \cite{deng2022nightlab}, BDD100K-night \cite{yu2018bdd100k} and Cityscapes \cite{cordts2016cityscapes}.
For all datasets, we use the standard mean Intersection over Union (mIoU) metric as an evaluation criterion. 

\subsection{Datasets}
\tx{NightCity \cite{tan2021night} is the first large-scale labeled night-time scene dataset for training and validation. NighCity+\cite{deng2022nightlab} refines some labeling errors in the validation set of NightCity \cite{tan2021night}.}
\tx{There is another night-time dataset, BDD100K-night, which selects  night-time images with their labels from the BDD100K \cite{yu2018bdd100k} as described in \cite{tan2021night} and \cite{deng2022nightlab}.}
\tx{Finally, we also test our model on a day-time dataset Cityscapes~\cite{cordts2016cityscapes} to verify its generalization ability}.

\textit{1) NightCity: }It includes 4,297 finely annotated images, of which 2,998 images are used for training, and 1,299 images are used for validation. 
The dataset labels are compatible with Cityscapes and contain 19 categories, and the resolution of the images is 512×1024. 

\textit{2) NightCity+: } NightCity+ updates the validation set of NightCity by correcting the labeling errors, and resizes the resolution of the image to 1024×2048.

\textit{3) BDD100K-night: }It has 320 images in the training set and 34 images in the validation set. It also has 19 categories same as Cityscapes and the image resolution is 720×1280.

\textit{4) Cityscapes: }It contains 5,000 annotated images, including 2,975 images for training, 500 images for validation, and 1,525 images for testing. 
The label contains 19 classes, and the resolution of the images is 1024×2048.

\subsection{Implementation Details}
The PyTorch framework is employed to implement our network. 
In the training phase, our model uses stochastic gradient descent (SGD) optimizer and a poly learning strategy with $\left(1-\frac{iter}{total\_iter}\right)^{0.9}$. 
We set the initial learning rate and weight decay coefficients to 5e-3 and 5e-4, respectively. 
Moreover, we set the batch size to 8, and the crop size is 384×768. We conduct experiments on one TITAN RTX GPU. 
For data augmentation, we use random scaling with ratio sampled in the range of (0.5, 2.0), random horizontal flip, crop, and Gaussian blur as in \cite{zhao2017pyramid}. 
And the training time is set to 260 epochs. For evaluation, we use multi-scale inference with ratios of [0.5, 0.75, 1.0, 1.25, 1.5, 1.75]. 
We utilize the dilated residual network\cite{he2016deep} as the backbone with an output stride of 1/8. 
In the process of SFF, to reduce the amount of computation, we use the projection function to reduce the number of channels to 512. 
We empirically set $\lambda _{1}=1$ and $\lambda _{2}=0.01$.



\subsection{Comparison on the NightCity and NightCity+}

To verify the effectiveness of our method, 
we train our model and other state-of-the-art methods on the NightCity train set and validate with both the NightCity validation set and the NightCity+ validation set, respectively. 
For experimental comparison consistency, we rescale the NightCity+ validation set images to 512×1024. 

\textit{Methods for Comparisons:}
To verify the effectiveness of our method, we compare our model with state-of-the-art methods including EGNet\cite{tan2021night} and NightLab\cite{deng2022nightlab} for Night-Time Scene Parsing (NTSP), PSPNet\cite{zhao2017pyramid}, DeeplabV3+\cite{chen2018encoder}, DANet\cite{fu2019dual}, CCNet\cite{huang2019ccnet}, GSCNN\cite{takikawa2019gated}, HANet\cite{choi2020cars}, STER\cite{zheng2021rethinking}, UperNet\cite{liu2021swin} and SegFormer\cite{xie2021segformer} for Day-Time Semantic Segmentation (DTSS). 
We report the performances of EGNet and HANet from \cite{tan2021night} and NightLab from \cite{deng2022nightlab}. 
PSPNet, Deeplabv3+, DANet and CCNet are trained with the same configurations as ours. 
Other methods use their official code and configurations for training.
Since our method can be applied to day-time segmentation methods for NTSP, we report the results of our model based on PSPNet\cite{zhao2017pyramid}, Deeplabv3+\cite{chen2018encoder} and CCNet\cite{huang2019ccnet}.

\begin{table*}[!t]
\centering
\caption{Comparison with state-of-the-arts on NightCity and NightCity+. 
DTSS represents day-time semantic segmentation, NTSP represents night-time scene parsing and MS stands for multi-scale inference. 
The Best results are highlighted in \textbf{bold}.\label{tab:ex_1}}
\renewcommand{\arraystretch}{1.3}
\begin{tabular}{l|c|c|c|c|c|c}
\hline
\multirow{2}{*}{\textbf{Method}}
& \multirow{2}{*}{\textbf{Years}}
& \multirow{2}{*}{\textbf{Original Task}}
& \multirow{2}{*}{\textbf{Backbone}}
& \multirow{2}{*}{\textbf{Resolution}}
& \multicolumn{2}{c}{\textbf{mIoU(\%)}} \\ 
\cline{6-7}
\multicolumn{1}{c}{} \vline
&\multicolumn{1}{c}{} \vline
&\multicolumn{1}{c}{} \vline
&\multicolumn{1}{c}{} \vline
&\multicolumn{1}{c}{} \vline
&\multicolumn{1}{c}{NightCity}\vline & {NightCity+} \\
\hline \hline
PSPNet\cite{zhao2017pyramid}    &CVPR 2017 &DTSS  &ResNet-101  &$512\times1024$        &51.02        &52.24  \\
DeeplabV3+\cite{chen2018encoder} &ECCV 2018 &DTSS  &ResNet-101  &$512\times1024$          &51.99       &53.26  \\
DANet\cite{fu2019dual}    &CVPR 2019 &DTSS   & ResNet-101   &$512\times1024$       &50.81       &52.47  \\
CCNet\cite{huang2019ccnet}    &ICCV 2019 &DTSS    &ResNet-101    &$512\times1024$       &49.81       &50.94  \\
GSCNN\cite{takikawa2019gated}    &ICCV 2019 &DTSS    &WideResNet38   &$512\times1024$        &48.92       &-  \\
HANet\cite{choi2020cars}    &CVPR 2020 &DTSS    &ResNet-101     &$512\times1024$      &51.1       &-  \\
STER \cite{zheng2021rethinking}    &CVPR 2021 &DTSS     &ViT-L  &$512\times1024$  & 43.11    & -  \\
UperNet\cite{liu2021swin}    &ICCV 2021 &DTSS      &Swin-T  &$512\times1024$  & 54.93    & -  \\
SegFormer\cite{xie2021segformer} &NeurIPS 2021 &DTSS      &MIT-B5  &$512\times1024$       & 46.28       & -  \\
EGNet\cite{tan2021night}  &TIP 2021 &NTSP  &ResNet-101 &$512\times1024$       & 51.8 & - \\     
NightLab (DeeplabV3+)\cite{deng2022nightlab}  &CVPR 2022 &NTSP     &ResNet-101 &$1024\times2048$       & - & 56.21   
\\ \hline
FDLNet (PSPNet)    &-  &NTSP   &ResNet-101 & $512\times1024$           &53.21   &54.25   \\ 
FDLNet (CCNet)    &-  &NTSP   &ResNet-101 & $512\times1024$           &51.00   &52.27   \\
FDLNet (DeeplabV3+)    &- &NTSP  &ResNet-101 & $512\times1024$           &54.60   &56.20    \\ 
FDLNet (DeeplabV3+) + MS    &-   &NTSP   &ResNet-101 & $512\times1024$  &\textbf{55.42}&\textbf{56.79}   \\ 
\hline
\end{tabular}
\end{table*}

\begin{table}[!t]
  \centering
  \caption{Comparison with state-of-the-arts on BDD100K-night.
  The best result is marked in \textbf{bold}.}
    \renewcommand{\arraystretch}{1.2}
    \begin{tabular}{l|c|c|c}
    \hline
    \textbf{Method} & \textbf{Years} & \textbf{Backbone} & \textbf{mIoU(\%)} \\   \hline \hline
    PSPNet\cite{zhao2017pyramid}     &   CVPR 2017    & ResNet-101      &19.62  \\
    Deeplabv3+\cite{chen2018encoder}     & ECCV 2018      &  ResNet-101      &23.42  \\
    DANet\cite{fu2019dual}    &CVPR 2019      &  ResNet-101      &21.06  \\
    CCNet\cite{huang2019ccnet}     &CVPR 2019       & ResNet-101      &17.74  \\
    SegFormer\cite{xie2021segformer}     & NeurIPS 2021 & MIT-B5      &22.06  \\
    AGLN\cite{li2022attention}     & TIP 2022      & ResNet-101      &20.16  \\
    \hline
    FDLNet (PSPNet)     & -  & ResNet-101      &25.00  \\
    FDLNet (CCNet)     & -  & ResNet101      &23.09  \\
    FDLNet (Deeplabv3+) & -  & ResNet-101      &\textbf{26.46}  \\
    \hline
    \end{tabular}%
  \label{tab:ex_2}%
\end{table}%

\textit{Quantitative Comparison:}
From Table \ref{tab:ex_1}, we can see that the day-time methods cannot achieve satisfying results due to the large gap between day and night scenes, but our proposed method can successfully adapt the day-time model to the night-time scenes.
Furthermore, our model based on the PSPNet\cite{zhao2017pyramid} outperforms the EGNet with a margin of 1.41\%. 
To gain better results, we utilize our model on a stronger baseline DeeplabV3+ \cite{chen2018encoder} and obtain 1.39\% improvement on NightCity and 1.95\% improvement on NightCity+. 
We also use a multi-scale strategy during inference and achieve a performance of 56.79\%, which outperforms the NightLab based on DeeplabV3+ with a margin of 0.58\%. Noting that the resolution is different between ours and NightLab.
Our method requires smaller resolution inputs which reduces computation but achieves higher performance.
The results show that our model improves the day-time models to adapt to NTSP and shows superior performance.

\begin{figure*}[!t]
    \centering
     \begin{minipage}{0.16\linewidth}
      \includegraphics[width=3.3cm]{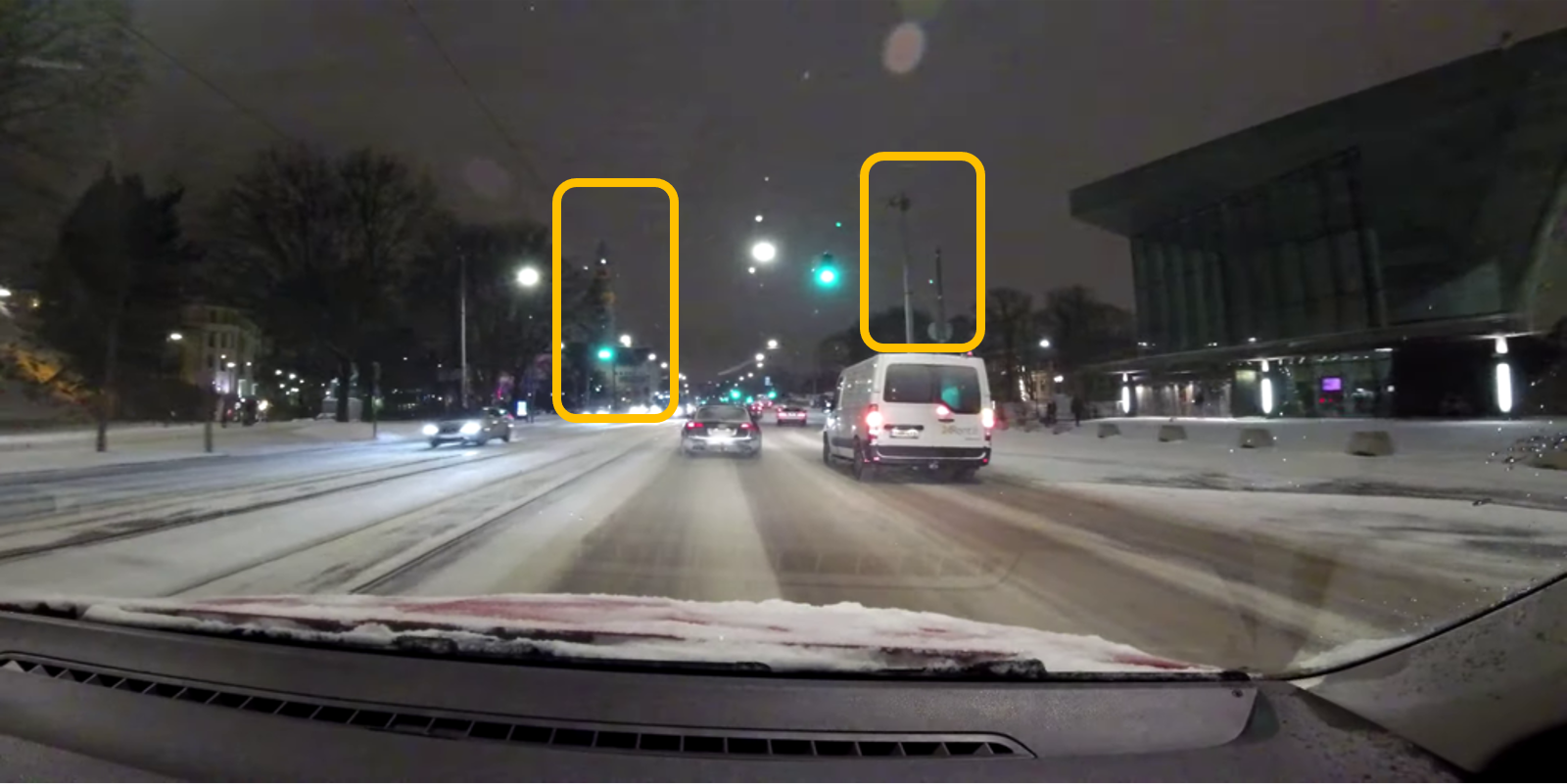}
    \end{minipage}
    \quad
    \begin{minipage}{0.16\linewidth}
      \includegraphics[width=3.3cm]{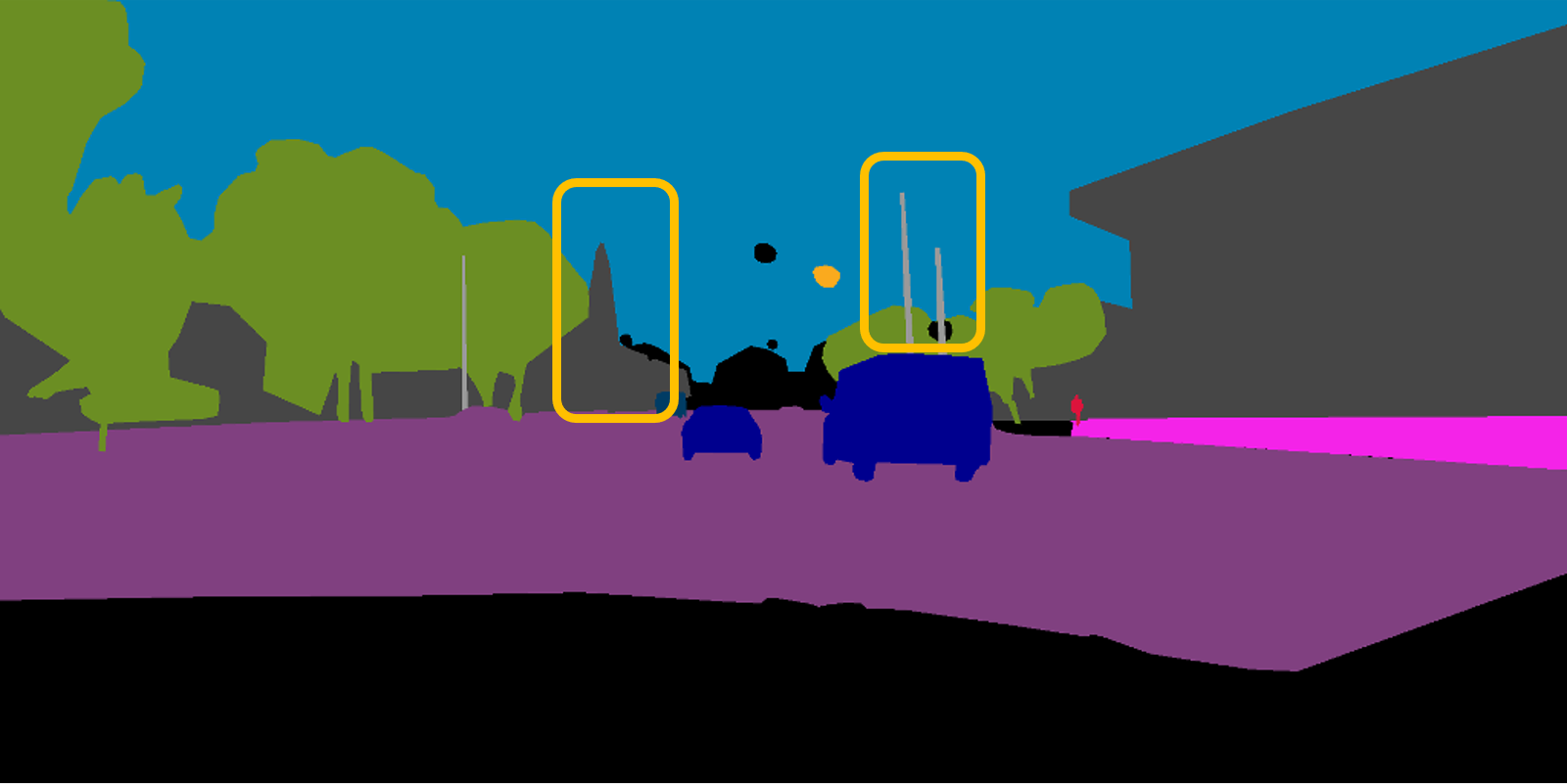}
    \end{minipage}
    \quad
    \begin{minipage}{0.16\linewidth}
      \includegraphics[width=3.3cm]{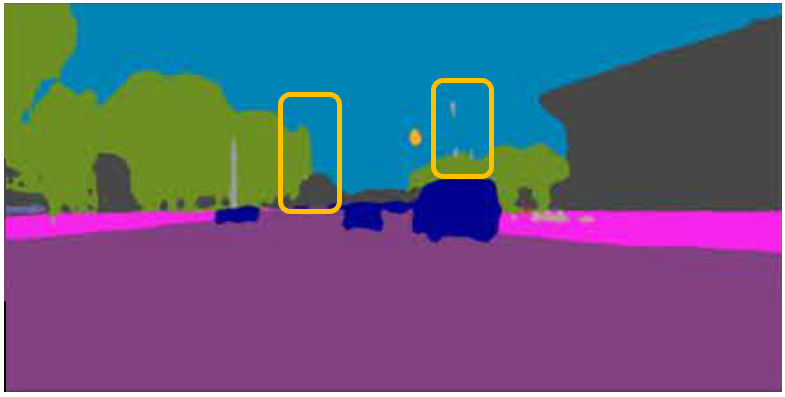}
    \end{minipage}
    \quad
    \begin{minipage}{0.16\linewidth}
      \includegraphics[width=3.3cm]{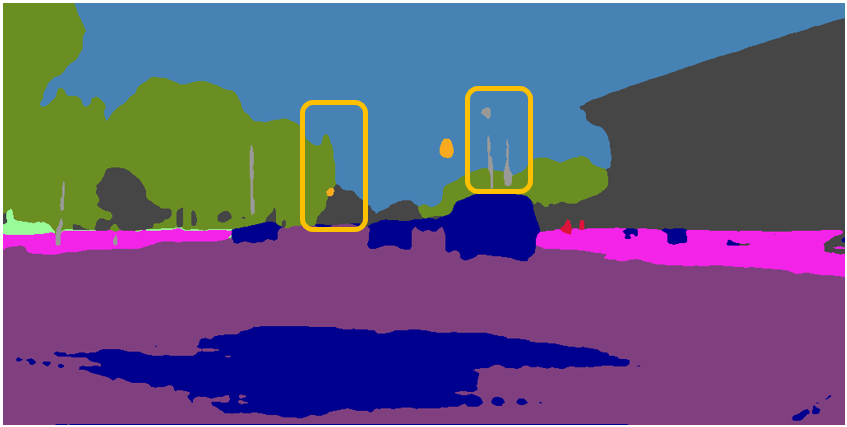}
    \end{minipage}
    \quad
    \begin{minipage}{0.16\linewidth}
      \includegraphics[width=3.3cm]{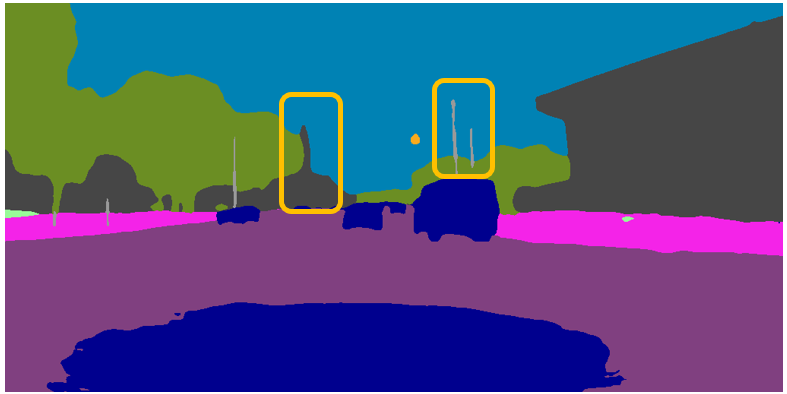}
    \end{minipage}
    
    
    \vskip 2pt
     \begin{minipage}{0.16\linewidth}
      \includegraphics[width=3.3cm]{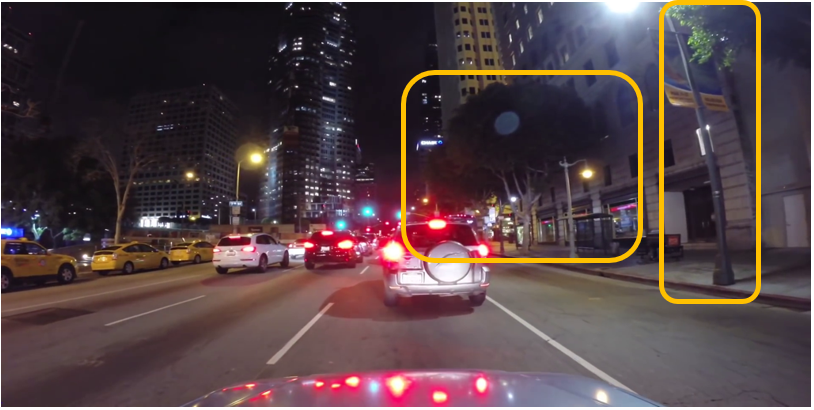}
    \end{minipage}
    \quad
    \begin{minipage}{0.16\linewidth}
      \includegraphics[width=3.3cm]{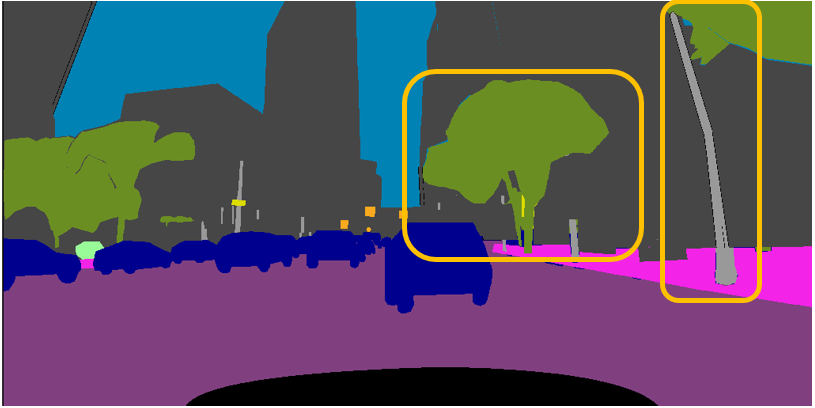}
    \end{minipage}
    \quad
    \begin{minipage}{0.16\linewidth}
      \includegraphics[width=3.3cm]{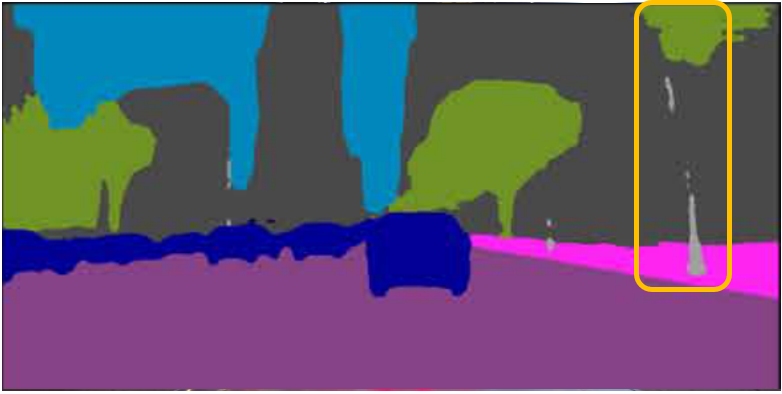}
    \end{minipage}
    \quad
    \begin{minipage}{0.16\linewidth}
      \includegraphics[width=3.3cm]{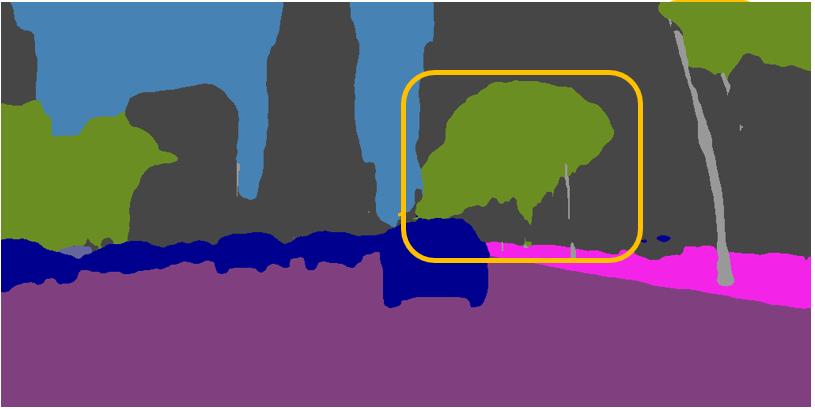}
    \end{minipage}
    \quad
    \begin{minipage}{0.16\linewidth}
      \includegraphics[width=3.3cm]{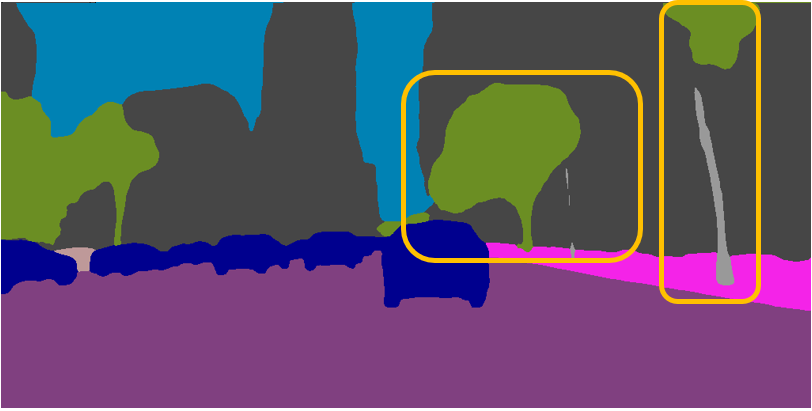}
    \end{minipage}
    
    \vskip 2pt
     \begin{minipage}{0.16\linewidth}
      \includegraphics[width=3.3cm]{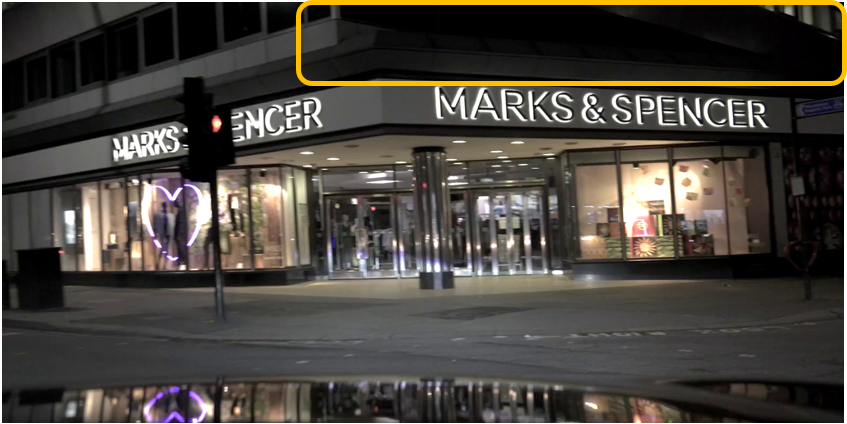}
    \end{minipage}
    \quad
    \begin{minipage}{0.16\linewidth}
      \includegraphics[width=3.3cm]{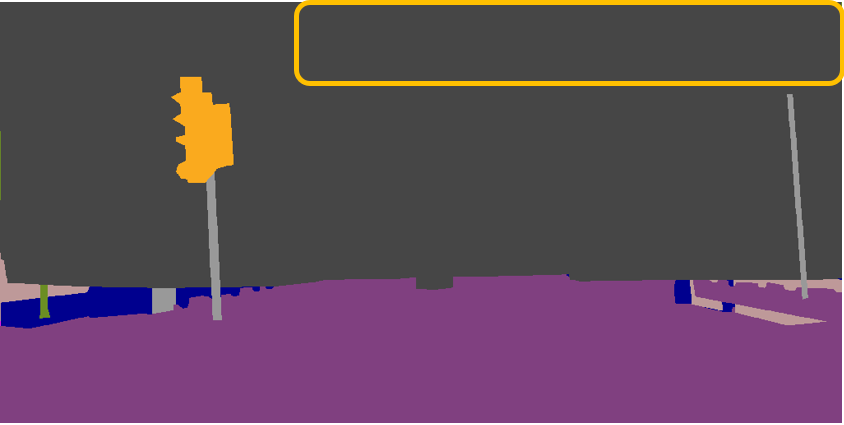}
    \end{minipage}
    \quad
    \begin{minipage}{0.16\linewidth}
      \includegraphics[width=3.3cm]{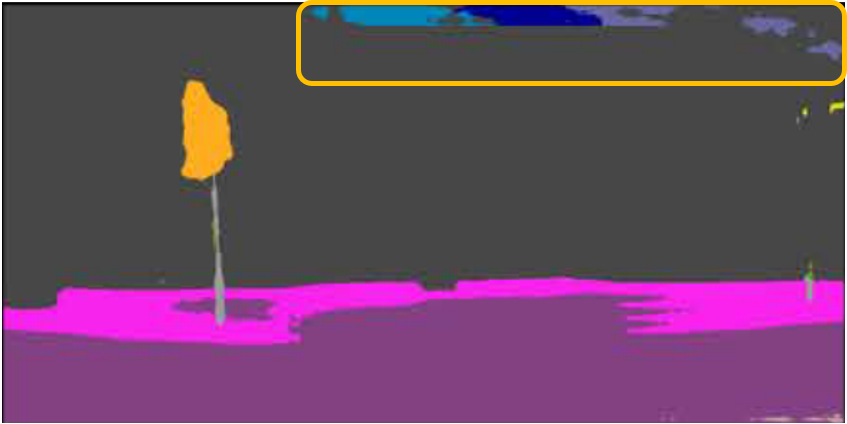}
    \end{minipage}
    \quad
    \begin{minipage}{0.16\linewidth}
      \includegraphics[width=3.3cm]{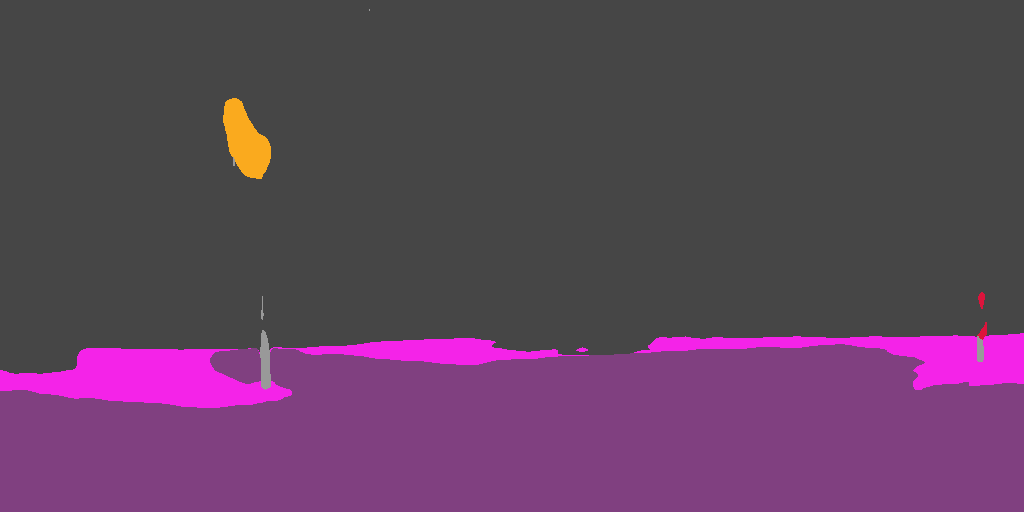}
    \end{minipage}
    \quad
    \begin{minipage}{0.16\linewidth}
      \includegraphics[width=3.3cm]{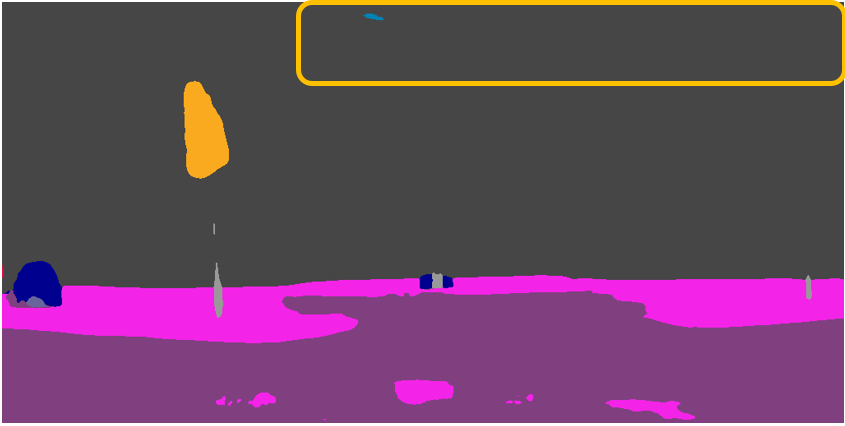}
    \end{minipage}
    
    \vskip 2pt
    \begin{minipage}{0.16\linewidth}
      \includegraphics[width=3.3cm]{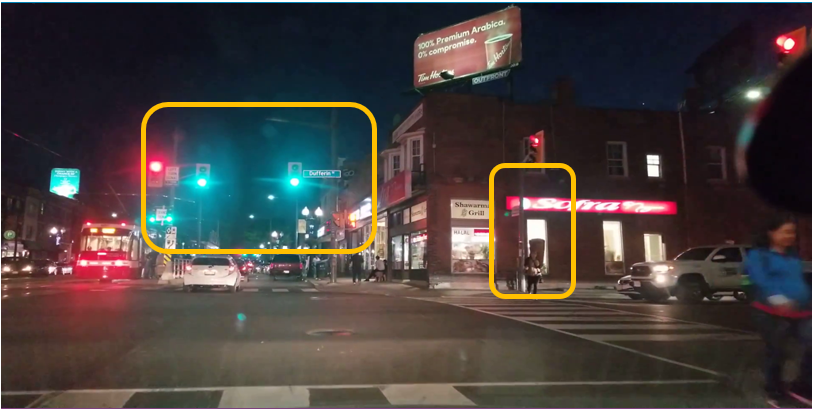}
      \centerline{Input}
    \end{minipage}
    \quad
    \begin{minipage}{0.16\linewidth}
      \includegraphics[width=3.3cm]{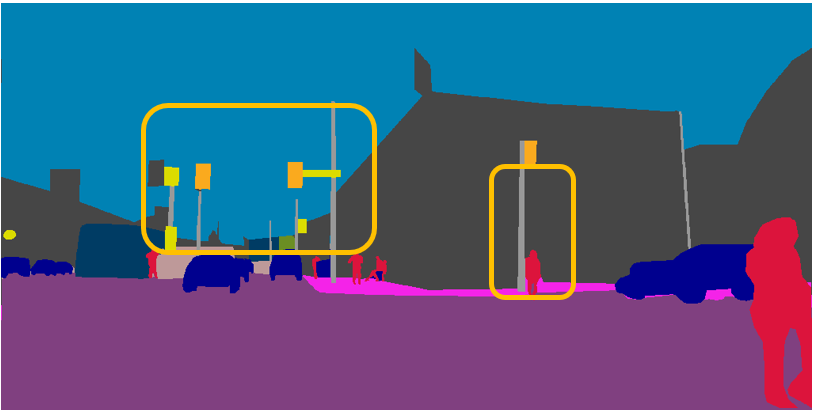}
      \centerline{GT}
    \end{minipage}
    \quad
    \begin{minipage}{0.16\linewidth}
      \includegraphics[width=3.3cm]{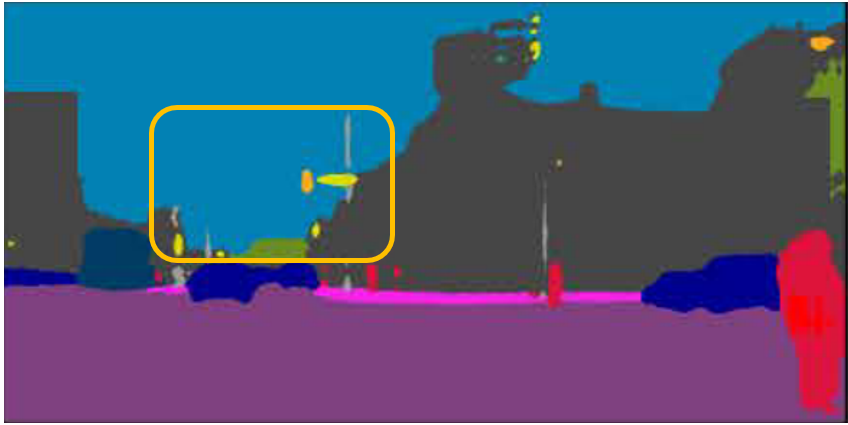}
      \centerline{EGNet\cite{tan2021night}}
    \end{minipage}
    \quad
    \begin{minipage}{0.16\linewidth}
      \includegraphics[width=3.3cm]{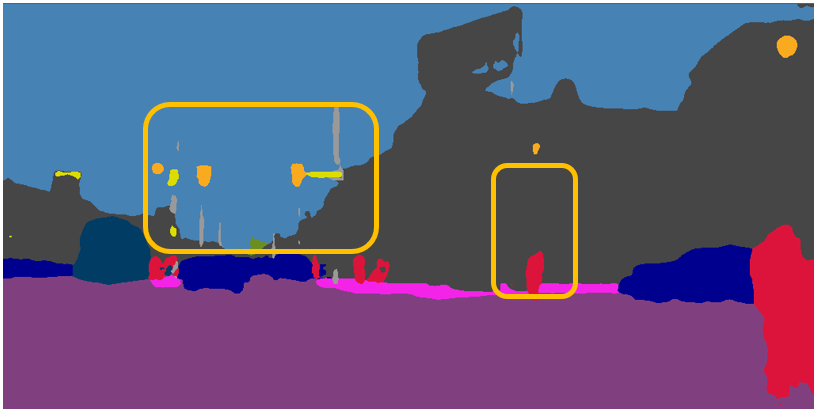}
      \centerline{UperNet\cite{liu2021swin}}
    \end{minipage}
    \quad
    \begin{minipage}{0.16\linewidth}
      \includegraphics[width=3.3cm]{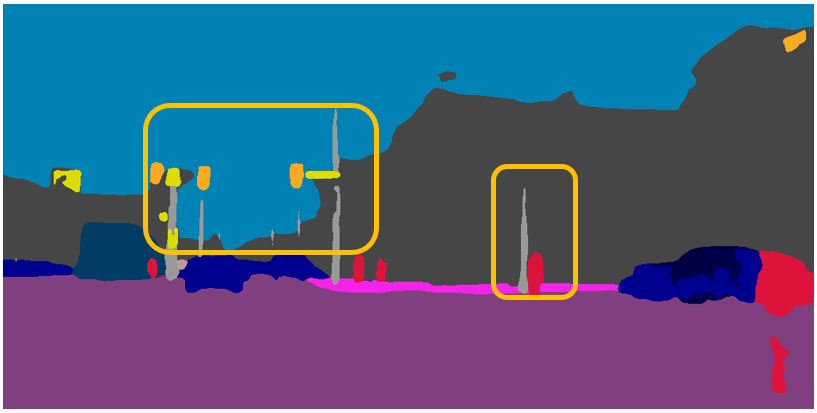}
      \centerline{Ours}
    \end{minipage}

\caption{Qualitative comparison on NightCity. Our advantages are highlighted by orange boxes.}
\label{fig:ex_1}
\end{figure*}

\textit{Qualitative Comparison:}
Figure \ref{fig:ex_1} quantitatively compares the prediction results of our model with state-of-the-art methods for DTSS and NTSP. 
Since NightLab does not show segmentation results on NightCity, we compare our results with EGNet and UperNet. 
The lighting conditions of night-time images make the frequency distribution quite different. 
Adjusting the lighting condition cannot allow the network to learn the frequency information, which makes the segmentation effect unsatisfactory. 
However, our model can handle this problem well. 
Particularly, in the first row, our model can identify areas of detail, such as distant buildings and poles. 
In the second row, our model gives more complete poles than EGNet and more complete trees than UperNet. 
In the third row, our model segments a more complete building. 
In the last row, EGNet cannot segment objects such as traffic lights, traffic signs and poles due to overexposure. 
UperNet cannot segment these objects completely.
But our model can recognize small objects with complete fineness. 
These results demonstrate the superior performance of our proposed model on NTSP.

\subsection{Comparison on the BDD100K-night}
We also conducted experiments on another labeled night-time scene dataset, BDD100K-night, to verify the effectiveness of our method.
We compare our model with state-of-the-art methods PSPNet\cite{zhao2017pyramid}, DeeplabV3+\cite{chen2018encoder}, DANet\cite{fu2019dual}, CCNet\cite{huang2019ccnet}, SegFormer\cite{xie2021segformer}, and AGLN\cite{li2022attention} for day-time semantic segmentation.
The results are reported in Table \ref{tab:ex_2}. 
We can see that our model based on DeeplabV3+ achieves the best performance of 26.46\%, which shows the generality of our proposed method.



\subsection{Model Analysis}

\textbf{Ablation Study.} To verify the effectiveness of different network components, we conduct five ablation studies.

\begin{figure}[ht]
    \centering
    \begin{minipage}{0.96\linewidth}
      \centerline{\includegraphics[width=7cm]{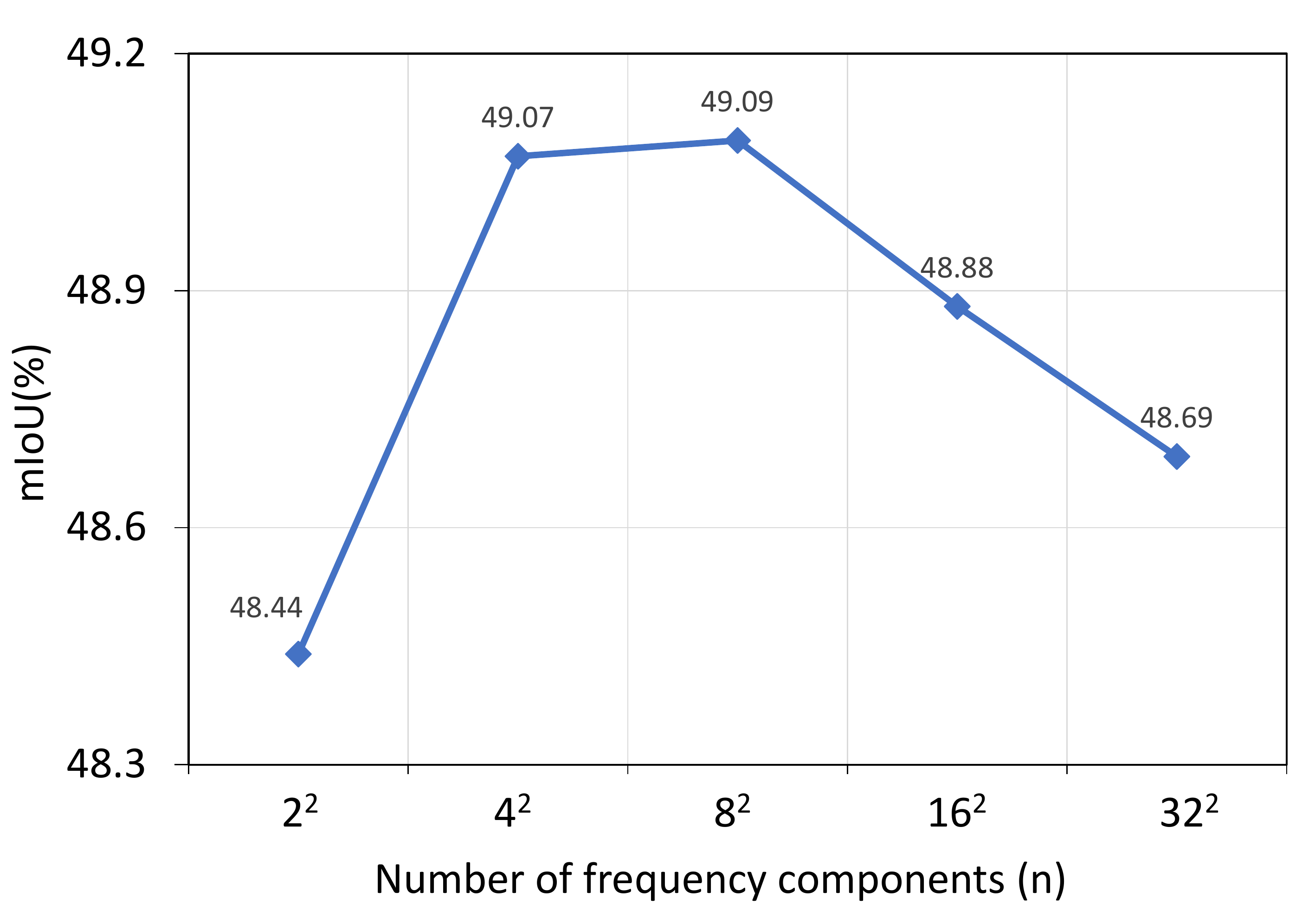}}
    \end{minipage}
    \vfill
\caption{Ablation Studies on the Number of Frequency Components.}
\label{fig:ab_1}
\end{figure}

\textit{1) Ablation Studies on the Number of Frequency Components:} 
The number of frequency components is one of the important factors affecting model performance. 
The network extracts image features and compresses the information into channel representations, so we use DCT to compress spatial features into N $\times$ N blocks to extract frequency features. 
Due to the limitation of channel numbers, N could be 2, 4, 8, 16, or 32. We use the ResNet-50 as the backbone and train the network for 120 epochs, as shown in Figure \ref{fig:ab_1}. 
We find that selecting 8 $\times$ 8 frequency components obtain the best performance.
In other experiments, we set the number of frequency components to 8 $\times$ 8.

\textit{2) Ablation Studies on Baseline Model:} We take the PSPNet\cite{zhao2017pyramid} as the baseline.
Due to the limitation of lighting conditions, night images have a lot of hard data\cite{deng2022nightlab}, so we use OHEM \cite{shrivastava2016training} during the training process to improve model performance. 
{In Table \ref{tab:ab_1}, dilated ResNet-101 is used as the backbone}, our baseline achieves 51.02\% and 0.88\% increase with OHEM ($2^{nd} $ row). 
A simply way to leverage frequency features is to design a SENet-like\cite{hu2018squeeze} module as in FcaNet\cite{qin2021fcanet}.
So we use the same way on LFE module to adjust the channel weights for the obtained frequency features and improve the performance from 51.90\% to 52.14\% ($3^{rd}$ row).
Then, we leverage SFF module in the network to replace the SENet-like module, in order to introduce the spatial context features extracted by PPM\cite{zhao2017pyramid}. The performance of the model is improved from 52.14\% to 52.84\% ($4^{th}$ row).
Incorporating the $L_{edge}$ brings about 0.37\% improvement ($5^{th}$ row), which shows the importance of edge supervision. 
The model further uses a multi-scale inference strategy (MS) to achieve a performance of 54.02\% ($6^{th}$ row).

\begin{table}[!t]
\centering
\caption{Ablation study. LFE stands for Learnable Frequency Encoder, SFF stands for Spatial Frequency Fusion Module, $L_{edge}$ stands for semantic edge loss, ohem stands for using Online Hard Example Mining during training. MS stands for Multi-Scale inference strategy\label{tab:ab_1}}
\renewcommand{\arraystretch}{1.2}
\begin{tabular}{l|cc|c}
\hline
\textbf{Method}       & \textbf{ohem}      & \boldmath{$L_{edge}$}\unboldmath  & \textbf{mIoU(\%)}    \\ \hline \hline
PSPNet      &   &      &  51.02              \\ 
             &$\surd$    &     &   51.90             \\ \hline
+ LFE      &$\surd$    &    &   52.14                    \\ \hline
+ LFE + SFF &$\surd$    &     &  52.84          \\
             &$\surd$    & $\surd$  & 53.21                \\ \hline
+ LFE + SFF + MS  &$\surd$    & $\surd$  & 54.02     \\  
\hline
\end{tabular}
\end{table}

\textit{3) Ablation Studies on Learnable Frequency Encoder:}  
To demonstrate the effectiveness of our proposed LFE module, we take PSPNet as a baseline and compare our method with two methods, one is using top-k components in \cite{qin2021fcanet} named (TOP), and the other is statically using all frequency components named (SA). 
We show the results in Table \ref{tab:ab_2}, our learnable frequency encoder strategy achieves the best performance improvement of 1.31\%.
However, we can see that the performance of TOP is better than SA, which indicates that simply leveraging all frequency components fails to adapt to NTSP due to the diverse frequency distribution.
Our method solves this issue by leveraging learnable frequency components.

\begin{table}[!t]
\centering
\caption{Ablation Studies on LFE. TOP stands for exploit TOP-k components, SA stands for Statically exploit all components and LFE stands for our proposed method Learnable Frequency Encoder. ohem stands for using Online Hard Example Mining during training.  \label{tab:ab_2}}
\renewcommand{\arraystretch}{1.2}
\begin{tabular}{l|c|c|c|c}
\hline
\textbf{Method} &\textbf{Backbone} &\textbf{ohem}   & \textbf{mIoU(\%)} &\boldmath$\Delta(\%)$\unboldmath  \\ \hline \hline
PSPNet  &ResNet-101    &$\surd$    & 51.90   &                    \\\hline
+ TOP   &ResNet-101    &$\surd$      & 52.67   & +0.77        \\ 
+ SA    &ResNet-101 &$\surd$   & 52.34   & +0.44                      \\
+ LFE  &ResNet-101  &$\surd$   & 53.21    & +1.31    \\
\hline
\end{tabular}
\end{table}

\begin{table}[!t]
\centering
\caption{Ablation Studies on SFF. FDLNet-SE stands for using SENet to replace our SFF module, $R_{s}$ is spatial representations and $\alpha$ is a scale parameter in SFF.\label{tab:ab_3}}
\renewcommand{\arraystretch}{1.2}
\begin{tabular}{l|c|c|c|c}
\hline
\textbf{Method} &\textbf{Backbone} &\textbf{ohem}   & \textbf{mIoU(\%)} &\boldmath$\Delta(\%)$\unboldmath\\ \hline \hline
FDLNet          &ResNet-101    &$\surd$ & 53.21  &                    \\\hline
FDLNet-SENet      &ResNet-101    &$\surd$  & 52.14   &-1.07        \\ 
w/o $R_{s}$     &ResNet-101    &$\surd$ & 52.06    &-1.15                     \\
w/o $\alpha$     &ResNet-101    &$\surd$ & 51.39   &-1.82                      \\
\hline
\end{tabular}
\end{table}

\begin{table*}[!t]
  \centering
  \caption{Improvements to Day-time Methods including performance comparison on three different validation sets and computation comparison.}
  \renewcommand{\arraystretch}{1.2}
    \begin{tabular}{l|c|c|c|c|c|c}
    \hline
    \textbf{Method} & \textbf{Backbone} & \textbf{Parameters} & \textbf{FLOPs} & \textbf{NightCity} & \textbf{NightCity+} & \textbf{BDD100K-night}\\\hline\hline
    PSPNet     & ResNet-101      &  70.12M     & 306.04G      & 51.02  &52.24 &19.62\\
    FDLNet (PSPNet)     & ResNet-101      &  71.83M     &  310.89G  &53.21 &54.25 &24.50\\ \hline
    DeeplabV3+   & ResNet-101      & 63.98M      &  314.02G     &51.99  &53.26   &23.42\\
    FDLNet (DeeplabV3+)      &  ResNet-101     & 67.46M     & 335.21G      &54.60  &56.20 &25.15\\ \hline
        CCNet     & ResNet-101      & 70.97M      & 329.55G      &49.81  &50.94 &17.74\\
    FDLNet (CCNet)     & ResNet-101      & 72.68M      & 334.41G      &51.00  &52.27 &21.82\\ 
    \hline
    \end{tabular}%
  \label{tab:ab_4}%
\end{table*}%

\textit{4) Ablation Studies on Spatial Frequency Fusion:}   
To verify the effectiveness of the SFF module, we conduct three experiments. 
(i) We use the structure of SENet\cite{hu2018squeeze} to replace the SFF module, which is named FDLNet-SENet. 
(ii) We only use frequency information to adjust the channel response without the aid of spatial information (w/o $R_{s}$). 
(iii) We verify the validity of the scale parameter alpha (w/o $\alpha$). 
As shown in Table \ref{tab:ab_3}, other alternative strategies degrade the model performance to varying degrees. 
Specifically, on the one hand, SENet leverages linear layers to adjust the channel response without the spatial features of the image, which can achieve good results in image classification but is not suitable for the NTSP task of spatial pixel-level classification. 
On the other hand, SFF leverages convolutional layers to adjust the channel response including the spatial structure information of the image, and achieves a 1.07\% improvement over SENet. 
The method that only utilizes frequency information has a similar structure to our SFF module, but its guiding effect is limited due to the lack of explicit spatial features. 
SFF utilizes features from two different domains (spatial and frequency) and outperforms the former by 1.15\%
The scale parameter $\alpha$ reduces gradient instability during training since adjusting the frequency component channel responses with information from all channels is a computationally expensive task. 
$\alpha$ can be changed incrementally to mitigate the drastic gradient changes, without the parameters, the performance of the model drops by 1.82\%.

\textit{5) Improvements to Day-time Methods:} Our method can be applied to the existing day-time methods to adapt them for the NTSP task. 
For consistency comparison, we modify PSP, DeeplabV3+, and CCNet by using our method with the same experimental settings. 
The results in Table \ref{tab:ab_4} show that our method improves the NTSP performance of existing day-time methods while introducing minimum computational overhead.

\begin{figure*}[!t]
    \centering
     \begin{minipage}{0.16\linewidth}
      \includegraphics[width=3.3cm]{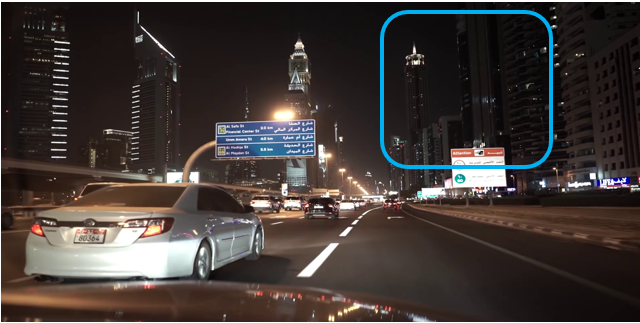}
    \end{minipage}
    \quad
    \begin{minipage}{0.16\linewidth}
      \includegraphics[width=3.3cm,]{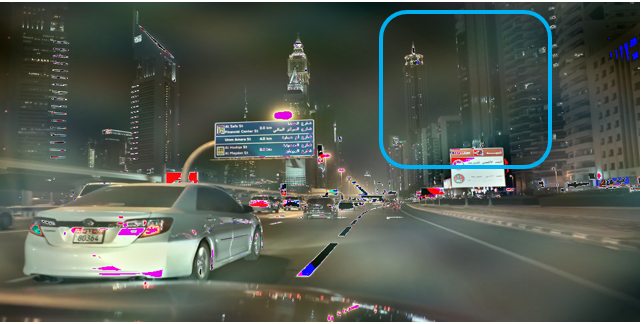}
    \end{minipage}
    \quad
    \begin{minipage}{0.16\linewidth}
      \includegraphics[width=3.3cm]{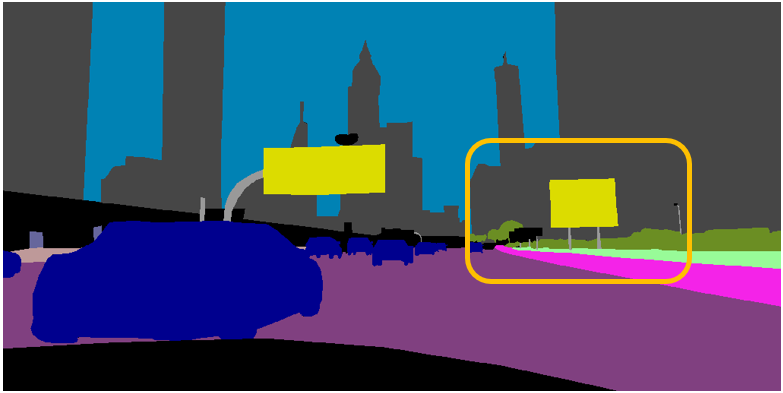}
    \end{minipage}
    \quad
    \begin{minipage}{0.16\linewidth}
      \includegraphics[width=3.3cm]{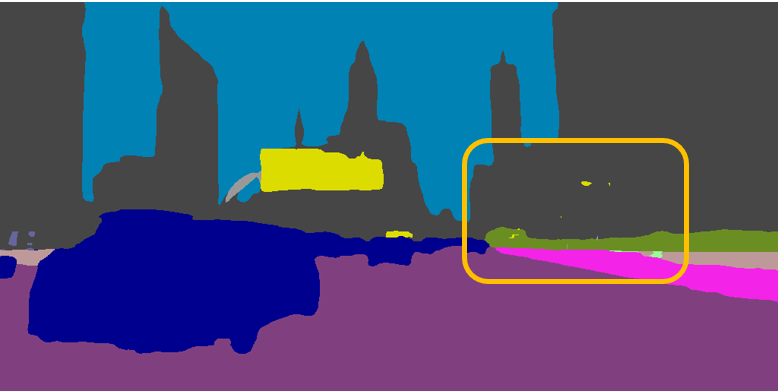}
    \end{minipage}
    \quad
    \begin{minipage}{0.16\linewidth}
      \includegraphics[width=3.3cm]{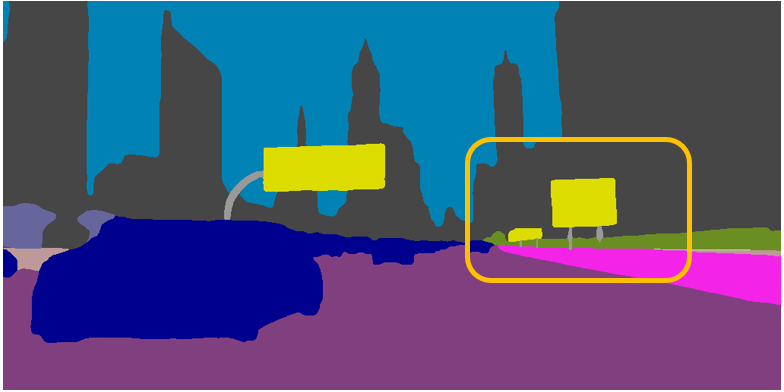}
    \end{minipage}
    
    \vskip 2pt
     \begin{minipage}{0.16\linewidth}
      \includegraphics[width=3.3cm]{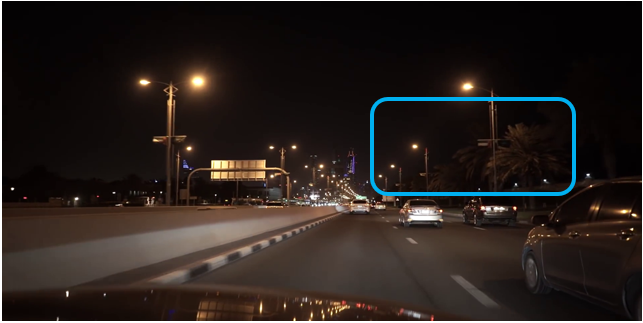}
    \end{minipage}
    \quad
    \begin{minipage}{0.16\linewidth}
      \includegraphics[width=3.3cm]{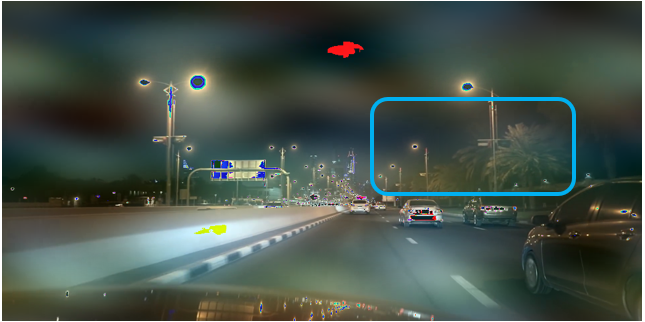}
    \end{minipage}
    \quad
    \begin{minipage}{0.16\linewidth}
      \includegraphics[width=3.3cm]{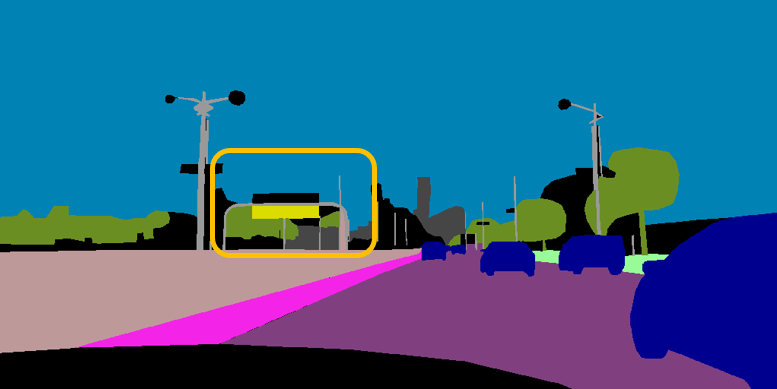}
    \end{minipage}
    \quad
    \begin{minipage}{0.16\linewidth}
      \includegraphics[width=3.3cm]{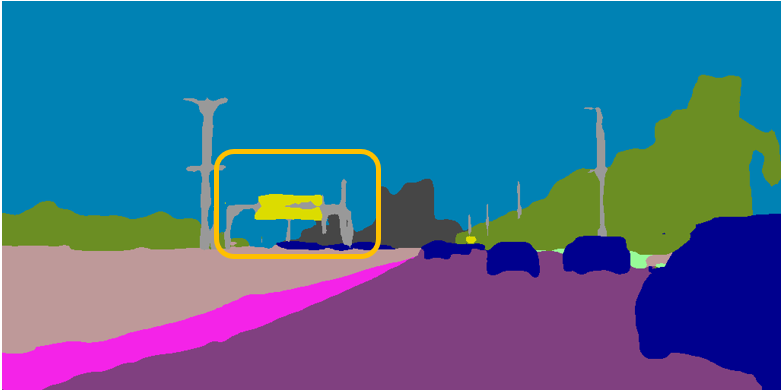}
    \end{minipage}
    \quad
    \begin{minipage}{0.16\linewidth}
      \includegraphics[width=3.3cm]{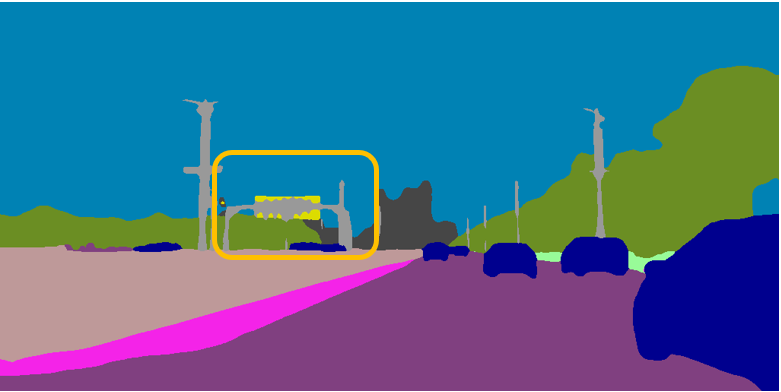}
    \end{minipage}
    
    
    \vskip 2pt
     \begin{minipage}{0.16\linewidth}
      \includegraphics[width=3.3cm]{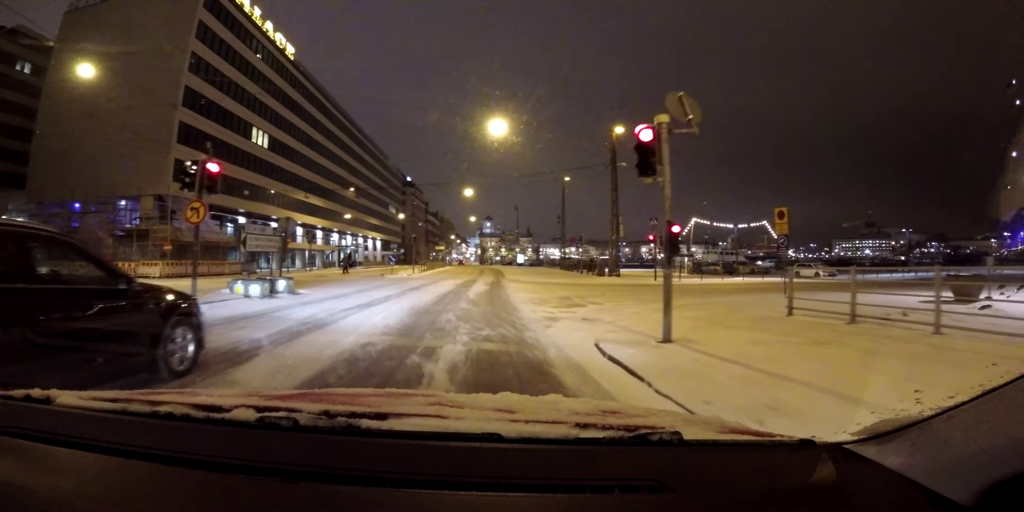}
    \end{minipage}
    \quad
    \begin{minipage}{0.16\linewidth}
      \includegraphics[width=3.3cm]{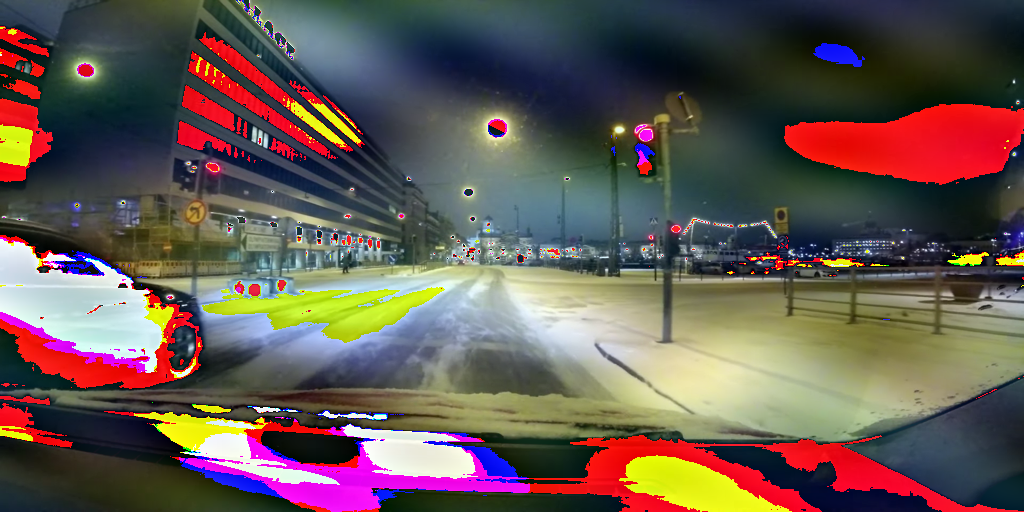}
    \end{minipage}
    \quad
    \begin{minipage}{0.16\linewidth}
      \includegraphics[width=3.3cm]{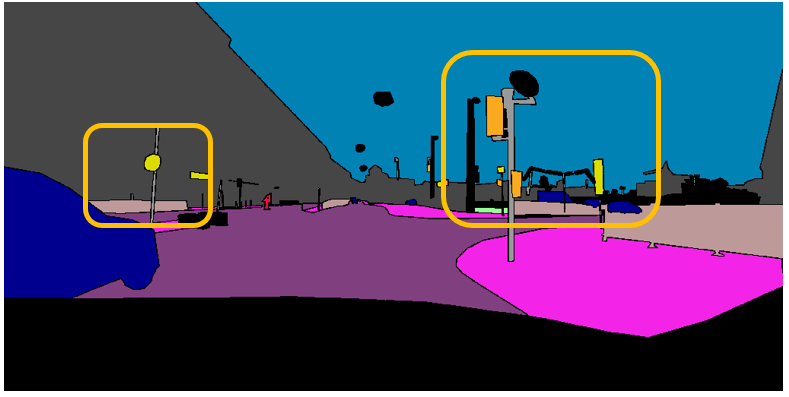}
    \end{minipage}
    \quad
    \begin{minipage}{0.16\linewidth}
      \includegraphics[width=3.3cm]{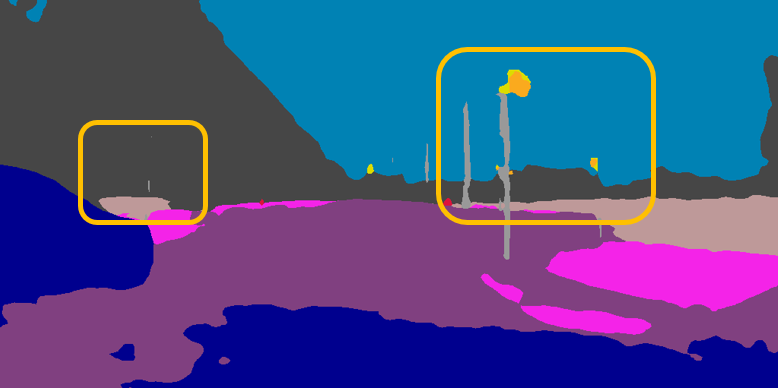}
    \end{minipage}
    \quad
    \begin{minipage}{0.16\linewidth}
      \includegraphics[width=3.3cm]{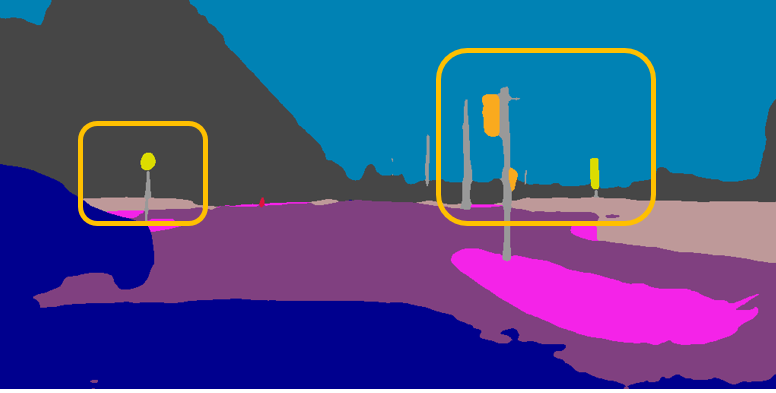}
    \end{minipage}
    
    \vskip 2pt
     \begin{minipage}{0.16\linewidth}
      \includegraphics[width=3.3cm]{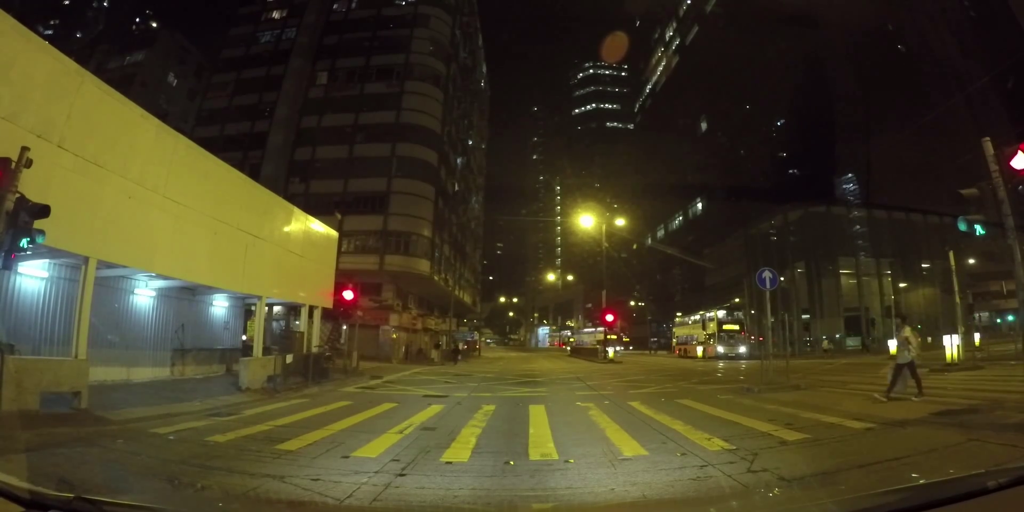}
    \end{minipage}
    \quad
    \begin{minipage}{0.16\linewidth}
      \includegraphics[width=3.3cm]{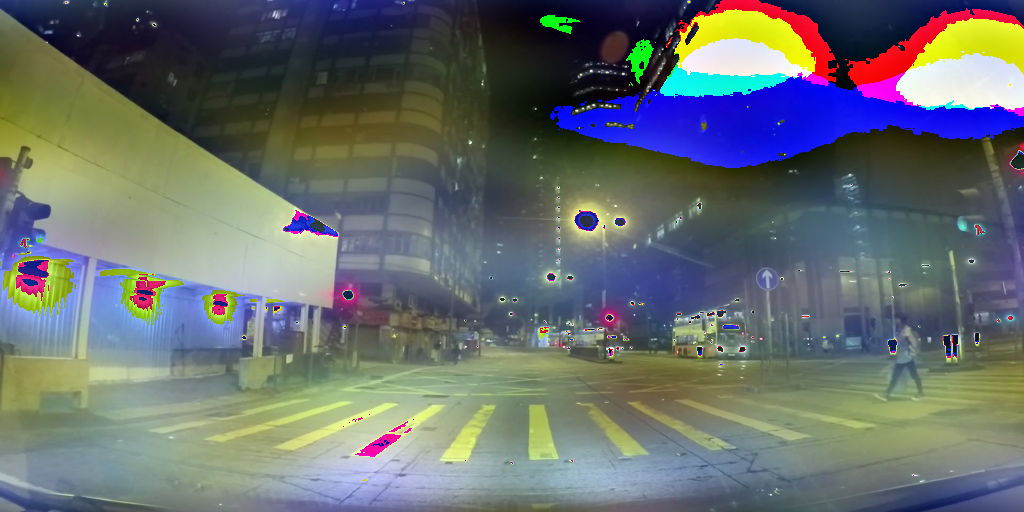}
    \end{minipage}
    \quad
    \begin{minipage}{0.16\linewidth}
      \includegraphics[width=3.3cm]{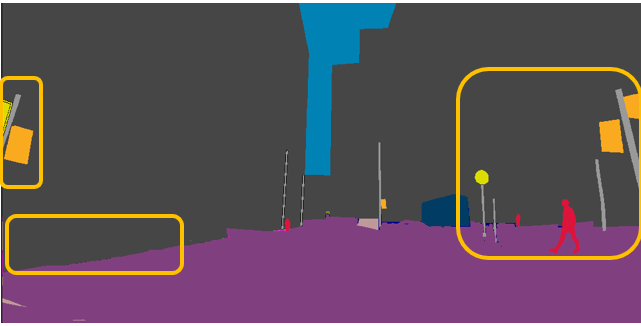}
    \end{minipage}
    \quad
    \begin{minipage}{0.16\linewidth}
      \includegraphics[width=3.3cm]{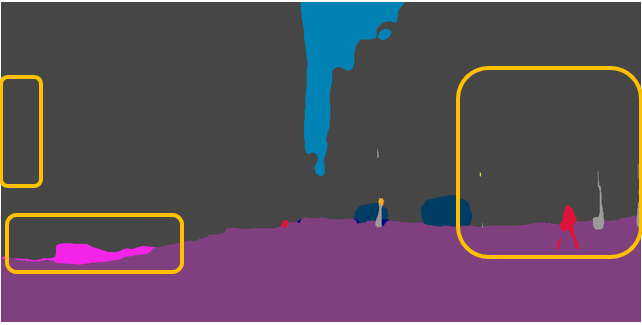}
    \end{minipage}
    \quad
    \begin{minipage}{0.16\linewidth}
      \includegraphics[width=3.3cm]{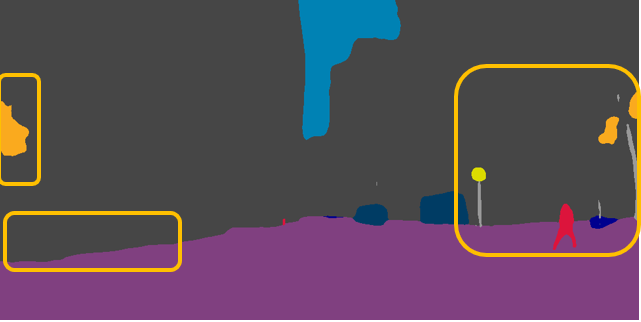}
    \end{minipage}
    
    
    \vskip 2pt
     \begin{minipage}{0.16\linewidth}
      \includegraphics[width=3.3cm]{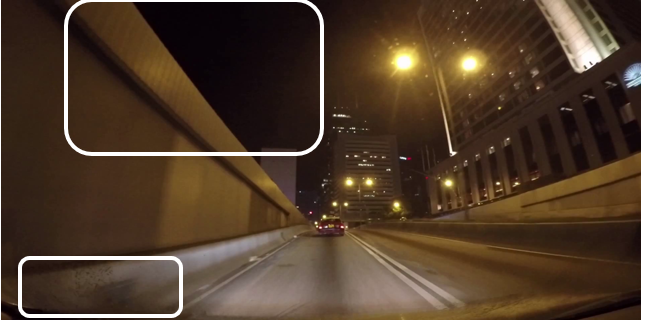}
    \end{minipage}
    \quad
    \begin{minipage}{0.16\linewidth}
      \includegraphics[width=3.3cm]{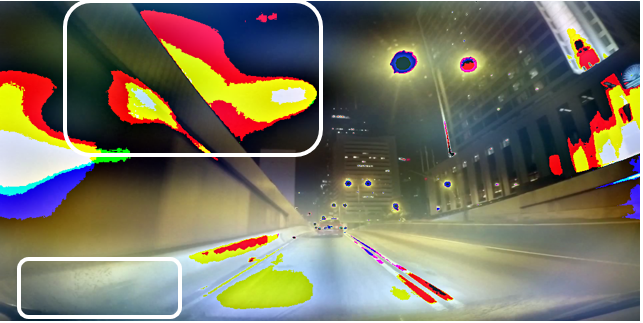}
    \end{minipage}
    \quad
    \begin{minipage}{0.16\linewidth}
      \includegraphics[width=3.3cm]{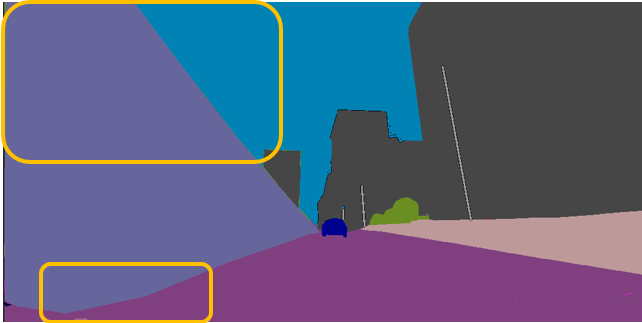}
    \end{minipage}
    \quad
    \begin{minipage}{0.16\linewidth}
      \includegraphics[width=3.3cm]{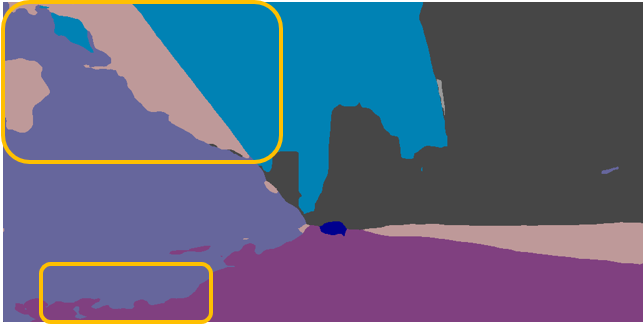}
    \end{minipage}
    \quad
    \begin{minipage}{0.16\linewidth}
      \includegraphics[width=3.3cm]{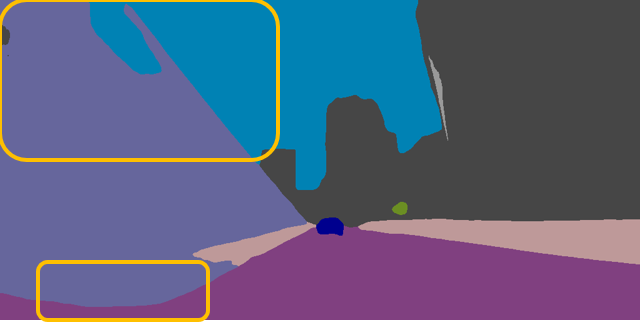}
    \end{minipage}
    
    \vskip 2pt
     \begin{minipage}{0.16\linewidth}
      \includegraphics[width=3.3cm]{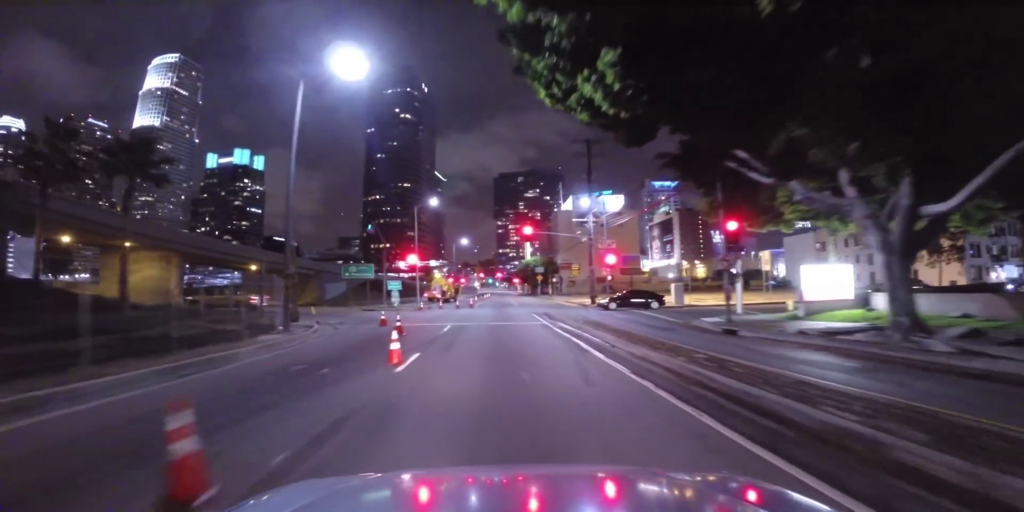}
    \end{minipage}
    \quad
    \begin{minipage}{0.16\linewidth}
      \includegraphics[width=3.3cm]{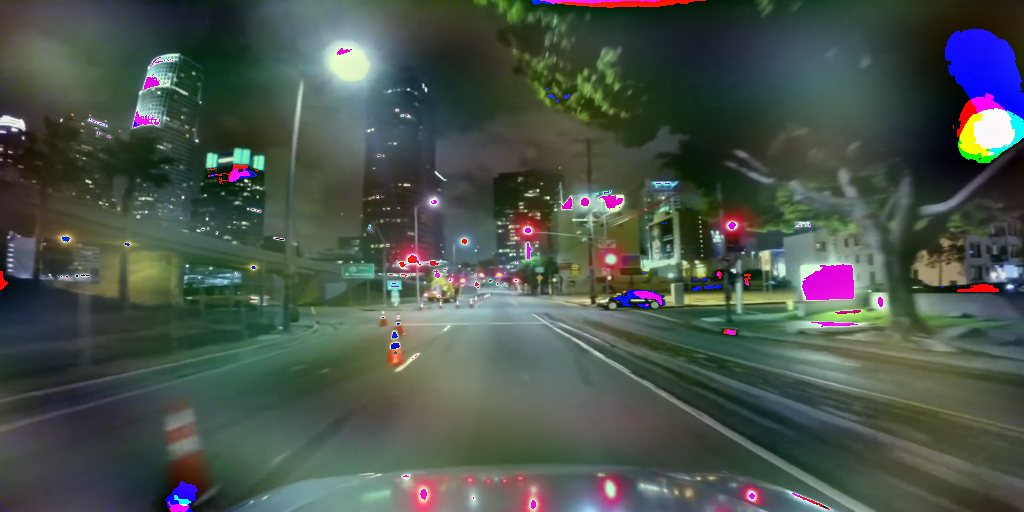}
    \end{minipage}
    \quad
    \begin{minipage}{0.16\linewidth}
      \includegraphics[width=3.3cm]{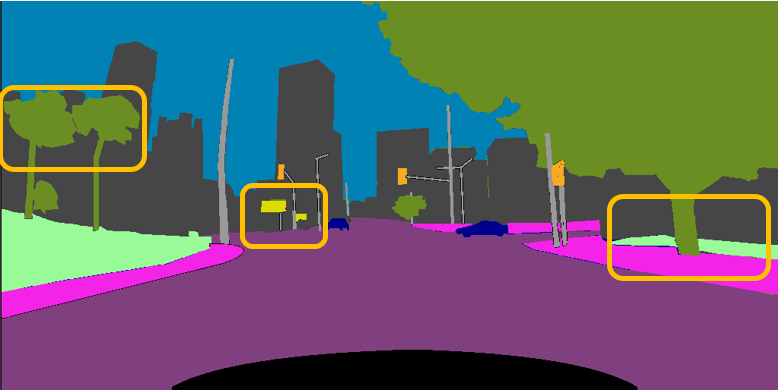}
    \end{minipage}
    \quad
    \begin{minipage}{0.16\linewidth}
      \includegraphics[width=3.3cm]{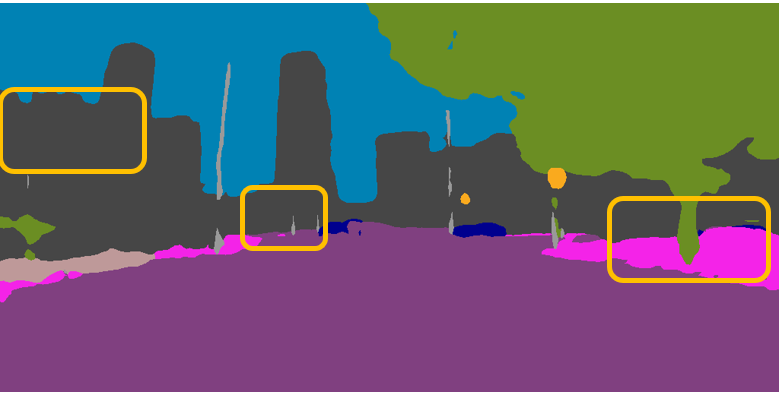}
    \end{minipage}
    \quad
    \begin{minipage}{0.16\linewidth}
      \includegraphics[width=3.3cm]{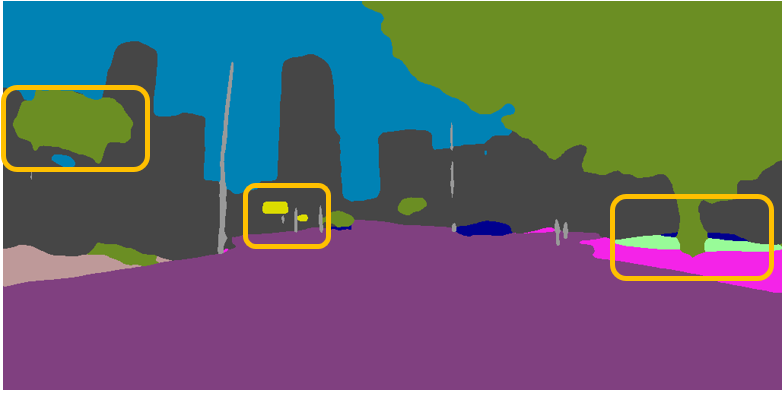}
    \end{minipage}
    
    \vskip 2pt
     \begin{minipage}{0.16\linewidth}
      \includegraphics[width=3.3cm]{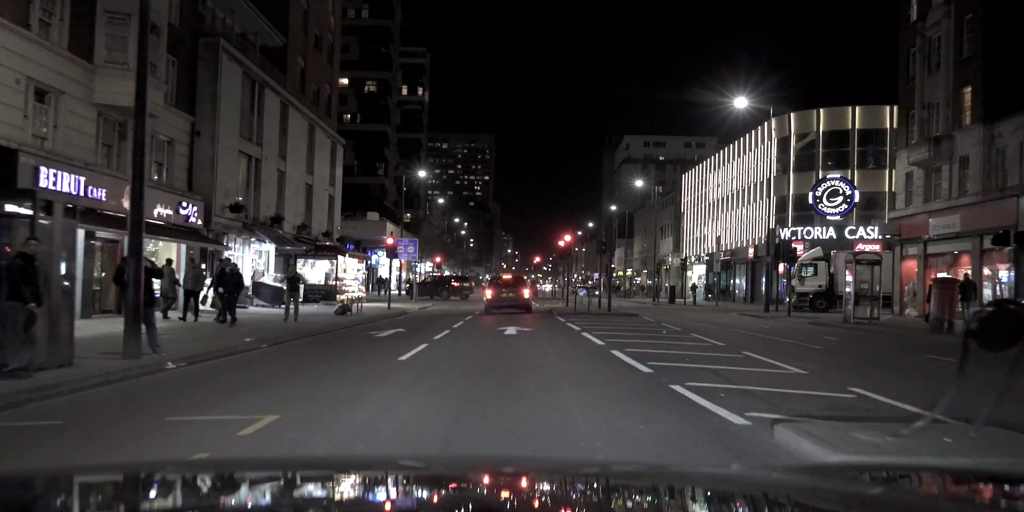}
    \end{minipage}
    \quad
    \begin{minipage}{0.16\linewidth}
      \includegraphics[width=3.3cm]{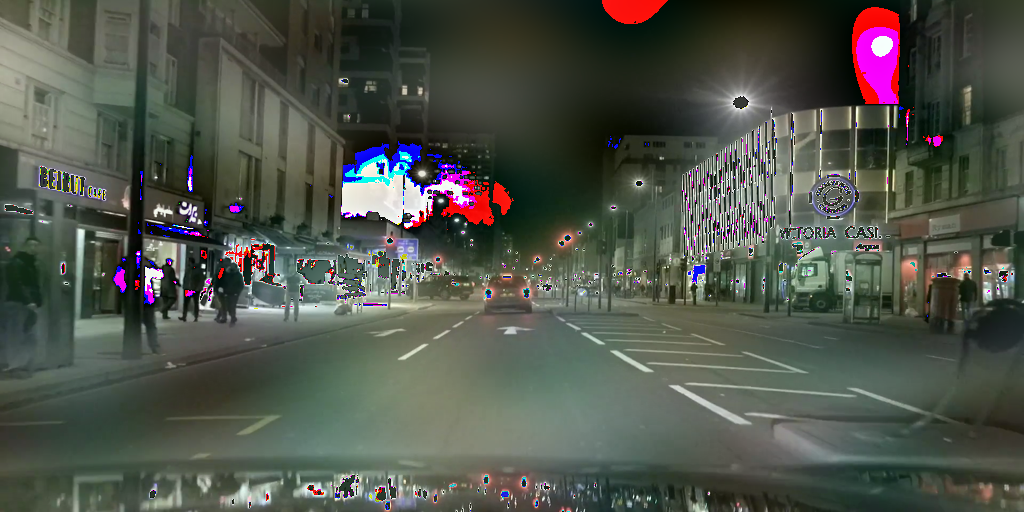}
    \end{minipage}
    \quad
    \begin{minipage}{0.16\linewidth}
      \includegraphics[width=3.3cm]{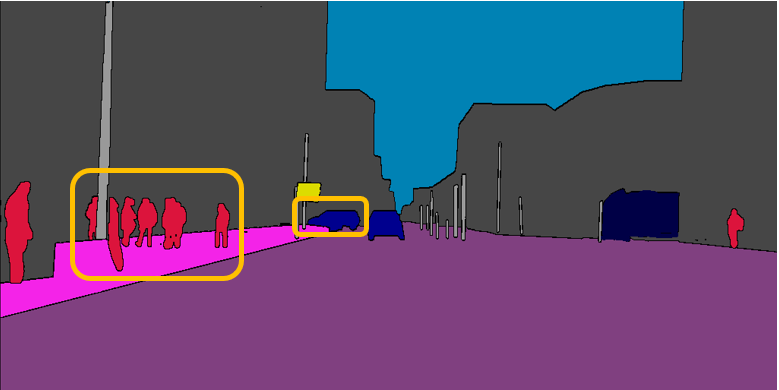}
    \end{minipage}
    \quad
    \begin{minipage}{0.16\linewidth}
      \includegraphics[width=3.3cm]{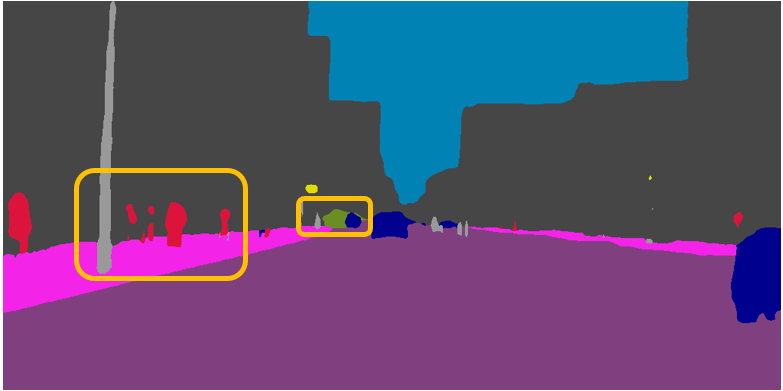}
    \end{minipage}
    \quad
    \begin{minipage}{0.16\linewidth}
      \includegraphics[width=3.3cm]{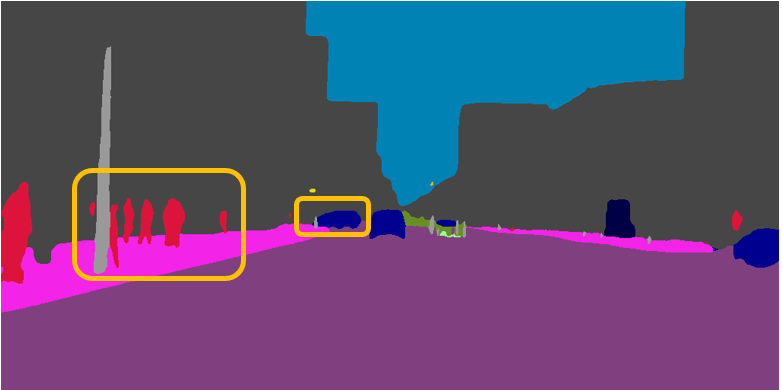}
    \end{minipage}
    
    \vskip 2pt
     \begin{minipage}{0.16\linewidth}
      \includegraphics[width=3.3cm]{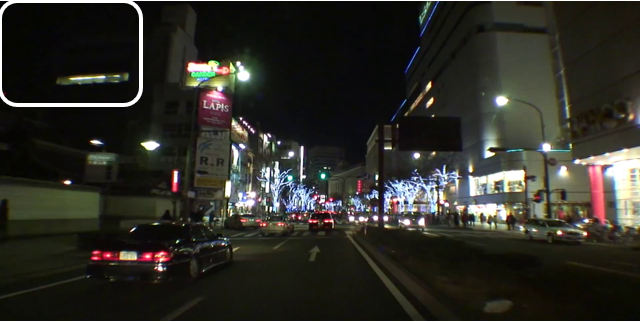}
    \end{minipage}
    \quad
    \begin{minipage}{0.16\linewidth}
      \includegraphics[width=3.3cm]{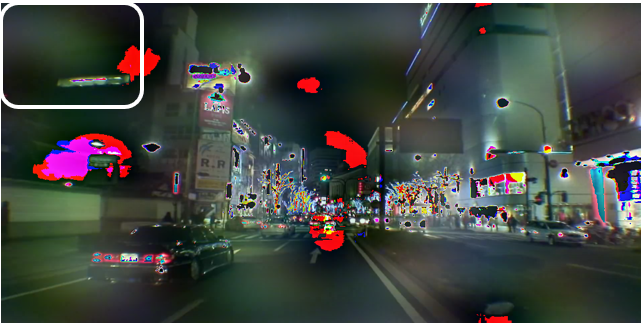}
    \end{minipage}
    \quad
    \begin{minipage}{0.16\linewidth}
      \includegraphics[width=3.3cm]{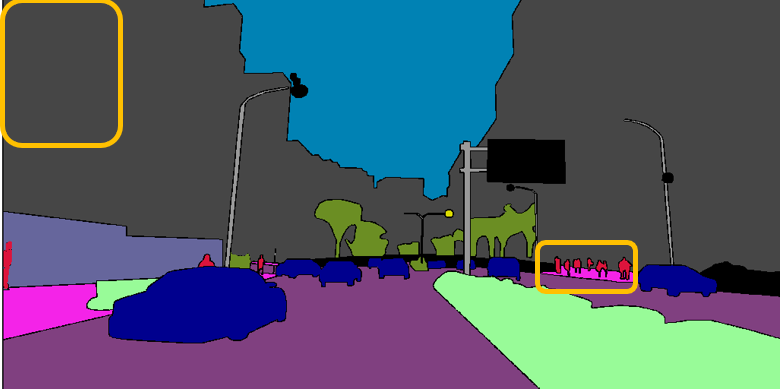}
    \end{minipage}
    \quad
    \begin{minipage}{0.16\linewidth}
      \includegraphics[width=3.3cm]{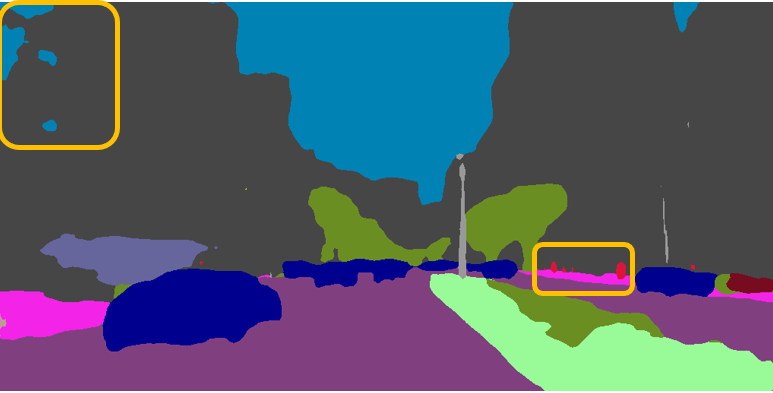}
    \end{minipage}
    \quad
    \begin{minipage}{0.16\linewidth}
      \includegraphics[width=3.3cm]{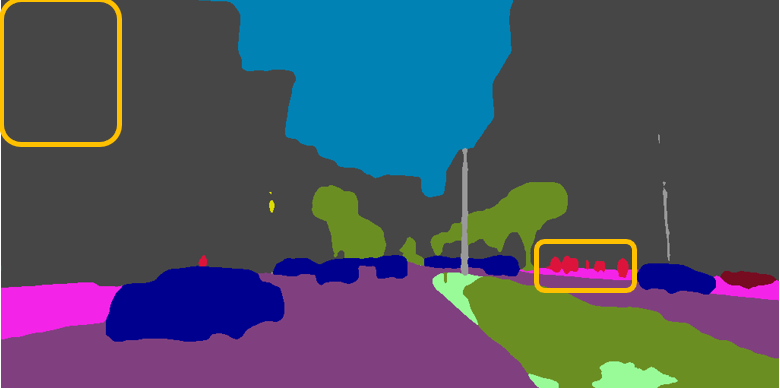}
    \end{minipage}

    \vskip 2pt
     \begin{minipage}{0.16\linewidth}
      \includegraphics[width=3.3cm]{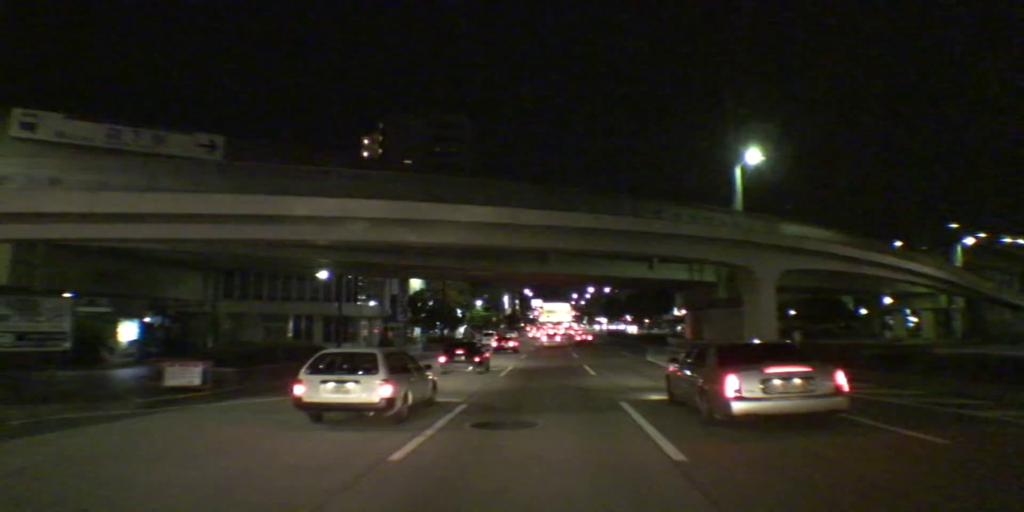}
      \centerline{Image}
    \end{minipage}
    \quad
    \begin{minipage}{0.16\linewidth}
      \includegraphics[width=3.3cm]{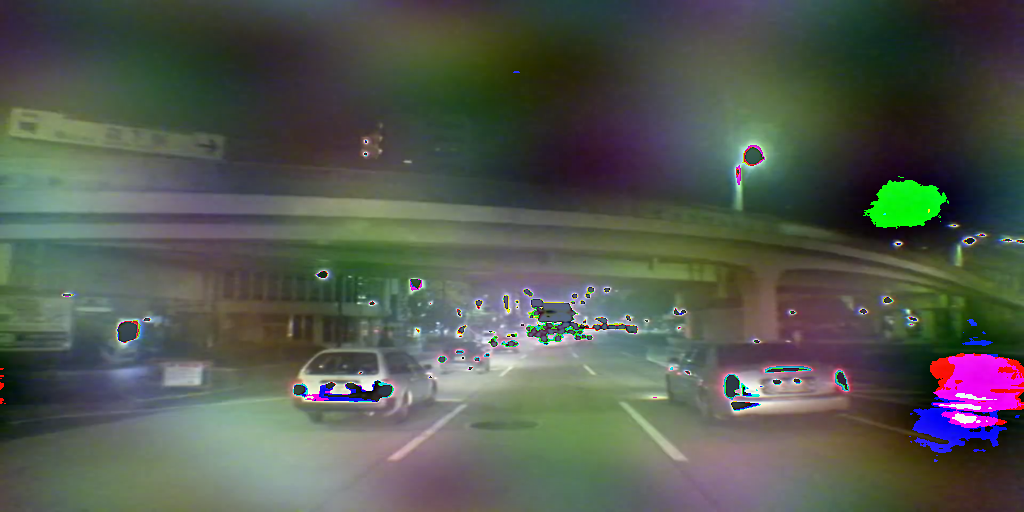}
      \centerline{FDA\cite{yang2020fda}}
    \end{minipage}
    \quad
    \begin{minipage}{0.16\linewidth}
      \includegraphics[width=3.3cm]{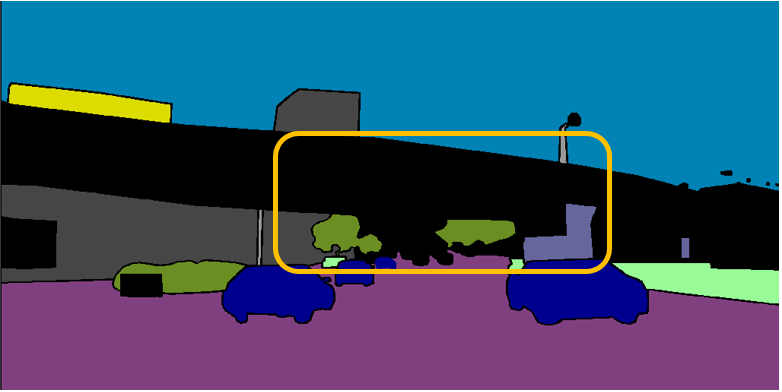}
      \centerline{GT}
    \end{minipage}
    \quad
    \begin{minipage}{0.16\linewidth}
      \includegraphics[width=3.3cm]{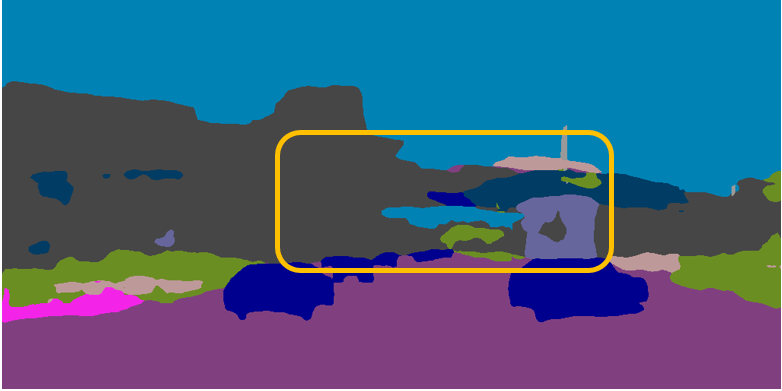}
      \centerline{FDA (DeeplabV3+)}
    \end{minipage}
    \quad
    \begin{minipage}{0.16\linewidth}
      \includegraphics[width=3.3cm]{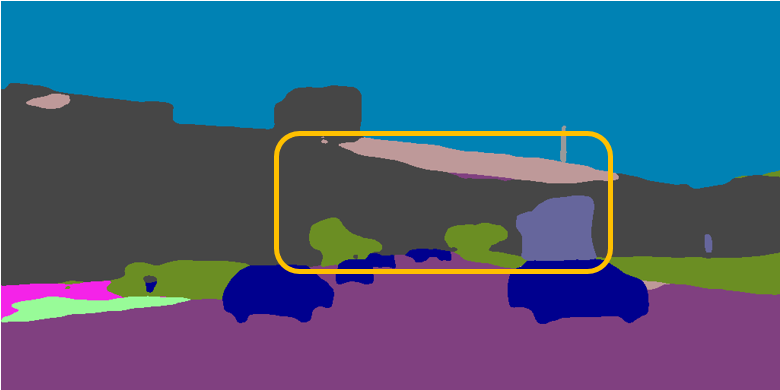}
      \centerline{Ours}
    \end{minipage}
    

\caption{Qualitative comparison on NightCity. 
The second column represents the style transformation of the image using Frequency Domain Adaption (FDA) \cite{yang2020fda}, and the fourth column (FDA (DeeplabV3+)) represents the prediction results on the style transformed dataset by using DeeplabV3+ \cite{chen2018encoder}. 
Successful cases of FDA are highlighted by blue boxes, and failure cases are highlighted by white boxes.
Our advantages are highlighted by orange boxes.}
\label{fig:ex_2}
\end{figure*}

\begin{table*}[!t]
  \centering
  \caption{Comparison with Frequency Domain Adaption (FDA) on NightCity. The method is evaluated on the resized NightCity+ validation set. The best results are marked in \textbf{bold}}
  \renewcommand{\arraystretch}{1.2}
  \resizebox{\linewidth}{!}{
    \begin{tabular}{c|ccccccccccccccccccc|c}
    \hline
    Method & road  & side. & bulid. & wall  & fence & pole  & light & sign  & vege. & terr. & sky   & pers. & rider & car   & truck & bus   & train & moto. & bicy. & mIoU \bigstrut\\
    \hline
    \hline
    DeeplabV3+     &90.4       &51.1       &83.2       &55.3       &53.5       &32.0       &24.4       &52.2       &59.0       &19.7       &88.2       &52.2       &25.2      &82.8       &\textbf{64.9}       &73.8       &59.1       &10.2       &34.7       &53.26 \bigstrut[t]\\
    FDA (DeeplabV3+)     &90.4       &50.6      &82.4       &53.5       &53.1       &30.1       &23.3       &49.0       &56.9       &20.6       &87.4       &49.6       &17.8       &82.3       &62.3       &72.4       &59.5       &0
    &34.1 &51.33\\ \hline

    FDLNet (PSPNet)     &90.5      &50.8       &83.2       &55.9       &53.1       &28.6       &24.8       &51.6       &59.1       &21.1       &87.9       &50.6       &25.2       &82.6       &63.1       &\textbf{75.1}       &\textbf{60.4}       &28.9       &38.3       &54.25\\
    FDLNet (DeeplabV3+)     &\textbf{91.2}       &\textbf{53.1}       &\textbf{83.8}       &\textbf{58.3}       &\textbf{54.4}       &\textbf{34.1}       &\textbf{30.1}       &\textbf{57.2}       &\textbf{60.1}       &\textbf{22.2}       &\textbf{88.2}       &\textbf{55.9}       &\textbf{27.6}       &\textbf{84.6}       &61.3       &73.8       &58.8       &\textbf{29.2}    
    &\textbf{44.0}    &\textbf{56.20}  \\
    \hline
    \end{tabular}
    }
  \label{tab:ex_3}
\end{table*}

\textbf{Comparisons with Frequency Domain Adaption.}
In some domain adaptation methods \cite{yang2020fda} \cite{xu2021cdada}, the frequency information is used to perform style transformation on the image to reduce the gap between the source and target domains, which is also suitable for the style transformation of day-time and night-time images. 
For comparison, we use NightCity as the source domain and Cityscapes as the target domain, so night-time images are transformed into daytime-like images by Frequency Domain Adaption (FDA)\cite{yang2020fda}. 
Then, we use DeeplabV3+ to obtain the prediction results. 
To gain more accurate results, we use the resized labels of the validation set of NightCity+ (512×1024) for comparison.

\textit{Qualitative Comparison:} 
We report the result in Figure \ref{fig:ex_2} and observe that transforming night-time images to day-time images using FDA can reduce the domain gap between them to a certain extent (blue boxes). 
However, simply replacing the frequency information of the two images often fails. 
For example, in the white boxes of the sixth and eighth rows, the transformed images are severely distorted, resulting in incomplete prediction results and chaotic boundaries, while our model uses learnable frequency information to guide the network to predict more complete predictions and the boundaries are clearer.
More comparison results are highlighted by orange boxes.
These visual results show that our proposed method is more efficient than directly preprocessing the image.

\textit{Quantitative Comparison:}
For better comparison, we also report the results of Deeplabv3+ on the original NightCity, as shown in Table \ref{tab:ex_3}. 
The performance of FDA (51.33\%) is even worse than the original method (53.26\%), which means that simply preprocessing the image with frequency information does not solve the NTSP problem well. 
Whereas our method introduces learnable frequency information into the model, the network learns the frequency distribution of night images and achieves better results.

\textbf{Compare with Day-time Dataset Cityscapes.}
Our method focuses on night-time scene parsing, because night-time scenes have two characteristics. 
First, the night scene contains information on all frequency components, including low-frequency areas with rich information and high-frequency areas with relatively little information. 
Moreover, the information contained in the high-frequency components of night-time images is richer than that of day-time images. 
Second, the frequency distribution of different night-time images is more different than day-time images, and the network can generate different component weights by learning each image. 

To explore the difference between our method on night-time and day-time images, we train models on Cityscapes and NightCity and perform quantitative comparisons on three different validation sets of Cityscapes, NightCity and NightCity+. 
The training settings are the same, except the learning rates are 0.005 and 0.01 for NightCity and Cityscapes, respectively. 
From Table \ref{tab:ab_9}, we can see that our method achieves better results than baselines on both day-time and night-time datasets. 
The improvement is 2.19\% on NighCity, 2.01\% on NighCity+ and 1.56\% on Cityscapes, which shows that our model is more effective on night-time images, and also shows the difference in frequency distribution between night-time and day-time images.
\begin{table}[!t]
  \centering
  \caption{Comparsion with the day-time dataset.}
    \begin{tabular}{l|c|c|c}
    \hline
    \textbf{Method} & \textbf{NighCity}& \textbf{NighCity+} & \textbf{Cityscapes} \\ \hline \hline
    PSPNet &51.02 &52.24       &70.86  \\
    FDLNet(PSPNet)&53.21  &54.25       &72.42 \\\hline
    $\Delta(\%)$    &+2.19   &+2.01       &+1.56  \\
  \hline
    \end{tabular}%
  \label{tab:ab_9}%
\end{table}%

\begin{figure}[htbp]
    \centering
     \subfloat[\rmfamily{Example 1}]{
            \centering
          \includegraphics[width=0.9\linewidth]{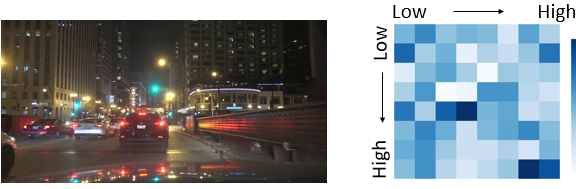}
          \label{fig:va_1}
    }
    \qquad 
     \subfloat[\rmfamily{Example 2}]{
            \centering
          \includegraphics[width=0.9\linewidth]{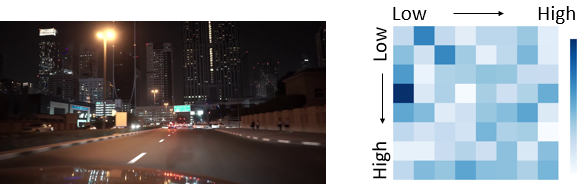}
          \label{fig:va_2}
    }
\caption{Image-level LFE heatmap. (left) The source image. (right) In the LFE heatmap, the low frequency components are located in the upper left part and the high frequency components are located in the lower right part. 
Different images correspond to different frequency affinities.}
\label{fig:VA_1}
\end{figure}

\begin{figure}[htbp]
    \centering
     \subfloat[\rmfamily{Ours}]{
            \centering
          \includegraphics[width=0.46\linewidth]{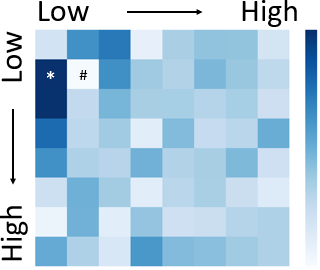}
    }
     \subfloat[\rmfamily{w/o $L_{edge}$}]{
            \centering
          \includegraphics[width=0.46\linewidth]{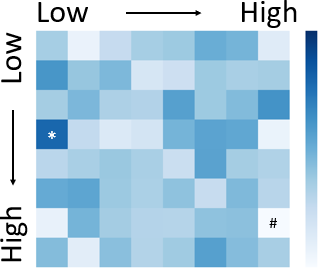}
    }
\caption{Dataset-level LFE heatmap. (left) Our model. (right) Our model w/o $L_{edge}$. $*$ represents the frequency component with the largest weight, and \# represents the frequency component with the smallest weight.}
\label{fig:VA_2}
\end{figure}

\subsection{Visual Analysis} 
\label{sec:VA}
To illustrate the capabilities of our proposed Learnable Frequency Encoder (LFE), we visualize the heatmap of LFE on the NightCity validation set.

\textit{1) Image-level LFE:} Our proposed LFE is able to dynamically adjust the weight of each frequency component, which means that there are differences of the frequency component affinity of each image. 
To illustrate this, we feed different images into the network to obtain the heatmap. 
Figure \ref{fig:VA_1} shows that for different images, the weights of frequency components are also different. 
The frequency components with the largest weight in Figure \ref{fig:va_1} appear in the high-frequency regions, but the distribution of the frequency weights is relatively loose. 
While the frequency component with the largest weight is located in the low-frequency regions as shown in Figure \ref{fig:va_2}, and the distribution of frequency weights is concentrated in low-frequency regions relatively.
This shows the frequency distribution in night-time scenes is diverse as we observe in Figure \ref{fig:intro_1} and Figure \ref{fig:intro_2}.

\textit{2) Dataset-level LFE:} To further analyze the LFE, we summed and averaged the LFE of all images in the validation set, resulting in a dataset-level LFE heatmap.
As shown in Figure \ref{fig:va_1}, we can see that our model prefers low-frequency components. 
The maximum weight of frequency component (*) is located in the low-frequency part, and the large weights are also generally concentrated in the low-frequency part, which proves that CNN prefers to select the low-frequency region with rich information in extracting features as \cite{xu2020learning} \cite{qin2021fcanet}. 
However, the minimum weight of frequency component (\#) is located in the low-frequency region rather than the high-frequency region, which reflects that the high-frequency information is equally important.

\textit{3) LFE on semantic edge loss:} Since we use the semantic edge loss $L_{edge}$, which focuses on the prediction of high-frequency related to semantic edges, so we visualize the LFE heatmap for analysis to demonstrate the effectiveness of semantic edge loss. 
Note that we use the same color numeric intervals in Figure \ref{fig:va_1} to visualize the results. 
Figure \ref{fig:va_2} shows the results w/o $L_{edge}$.
The maximum weight (*) is located in the low-frequency part, and the minimum (\#) is located in the high-frequency part.
In contrast to method w/ $L_{edge}$, whose minimum weight (\#) appears in the low-frequency region. 
This shows that semantic edge loss strengthens the attention of edge details to a certain extent. 
Furthermore, the model w/o $L_{edge}$ overall has a looser selection of frequency components compared to the model w/ $L_{edge}$, which indicates the semantic edge loss enforces the network to extract frequency features more efficiently and reduce information redundancy.

\section{Conclusion}
In this paper, we propose a Frequency Domain Learning Network (FDLNet) to handle the frequency information distribution diversification of Night-Time Scene Parsing (NTSP). 
Specifically, the Learnable Frequency Encoder (LFE) adjusts the weights of frequency components generated by the DCT. 
Since high and low-frequency information is both important for NTSP, the encoder processes all frequency components information. 
Moreover, the encoder dynamically adjusts each frequency component to adapt to changes in the frequency distribution of night images. 
Furthermore, the Spatial Frequency Fusion module (SFF) leverages information from two different domains to guide the network segmentation since only utilizing frequency information lacks spatial context features that are important for NTSP.
Besides, our method allows a simple modification of the day-time model to adapt it to night-time scenes. 
Our model achieves state-of-the-art performance on NightCity and competitive results on NightCity+ and BDD100K-night.

\section*{Acknowledgements}

This work is partially supported by the National Natural Science Foundation of China (No. 61972157),
Shanghai Municipal Science and Technology Major Project (2021SHZDZX0102),
Shanghai Science and Technology Commission (21511101200). 
We also appreciate the help of researchers other than the authors of the paper.

\bibliographystyle{ieeetr}
\bibliography{Mytex.bib}

\end{document}